\documentclass{article}

\usepackage{longtable}
\usepackage{booktabs}
\usepackage{appendix}
\usepackage{times}
\usepackage{subcaption}
\usepackage{xcolor}
\usepackage{algpseudocode}
\usepackage{algorithm}
\usepackage{amsmath}
\usepackage{amssymb}
\usepackage{array}
\usepackage{float}
\usepackage{floatflt}
\usepackage{graphicx}
\usepackage[bookmarks=true]{hyperref}
\usepackage{multicol}
\usepackage{wrapfig}
\usepackage{enumitem}

\setlist[itemize]{
    nosep,
    before=\vspace{-\parskip},
    after=\vspace{-\parskip},
    leftmargin=12pt
}

\usepackage[final]{corl_2025} 

\title{Sample-Efficient Online Control Policy Learning with Real-Time Recursive Model Updates}

\author{
  Zixin Zhang,~James Avtges,~Todd D. Murphey\\
  Department of Mechanical Engineering\\
  Northwestern University, United States\\
  \texttt{\{zixinz,~javtges\}@u.northwestern.edu,~t-murphey@northwestern.edu}
}

\begin{document}
\maketitle

\begin{abstract}

Data-driven control methods need to be sample-efficient and lightweight, especially when data acquisition and computational resources are limited---such as during learning on hardware. Most modern data-driven methods require large datasets and struggle with real-time updates of models, limiting their performance in dynamic environments. Koopman theory formally represents nonlinear systems as linear models over observables, and Koopman representations can be determined from data in an optimization-friendly setting with potentially rapid model updates. In this paper, we present a highly sample-efficient, Koopman-based learning pipeline: Recursive Koopman Learning (RKL). 
We identify sufficient conditions for model convergence and provide formal algorithmic analysis supporting our claim that RKL is lightweight and fast, with complexity independent of dataset size. We validate our method on a simulated planar two-link arm and a hybrid nonlinear hardware system with soft actuators, showing that real-time recursive Koopman model updates improve the sample efficiency and stability of data-driven controller synthesis---requiring only $<$10\% of the data compared to benchmarks. The high-performance C++ codebase is open-sourced. Website: \href{https://www.zixinatom990.com/home/robotics/corl-2025-recursive-koopman-learning}{https://www.zixinatom990.com/home/robotics/corl-2025-recursive-koopman-learning}.



\end{abstract}

\section{Introduction}\label{sec:introduction}

\begin{wrapfigure}[11]{r}{0.49\textwidth}
    \centering
    \includegraphics[width=0.49\textwidth]{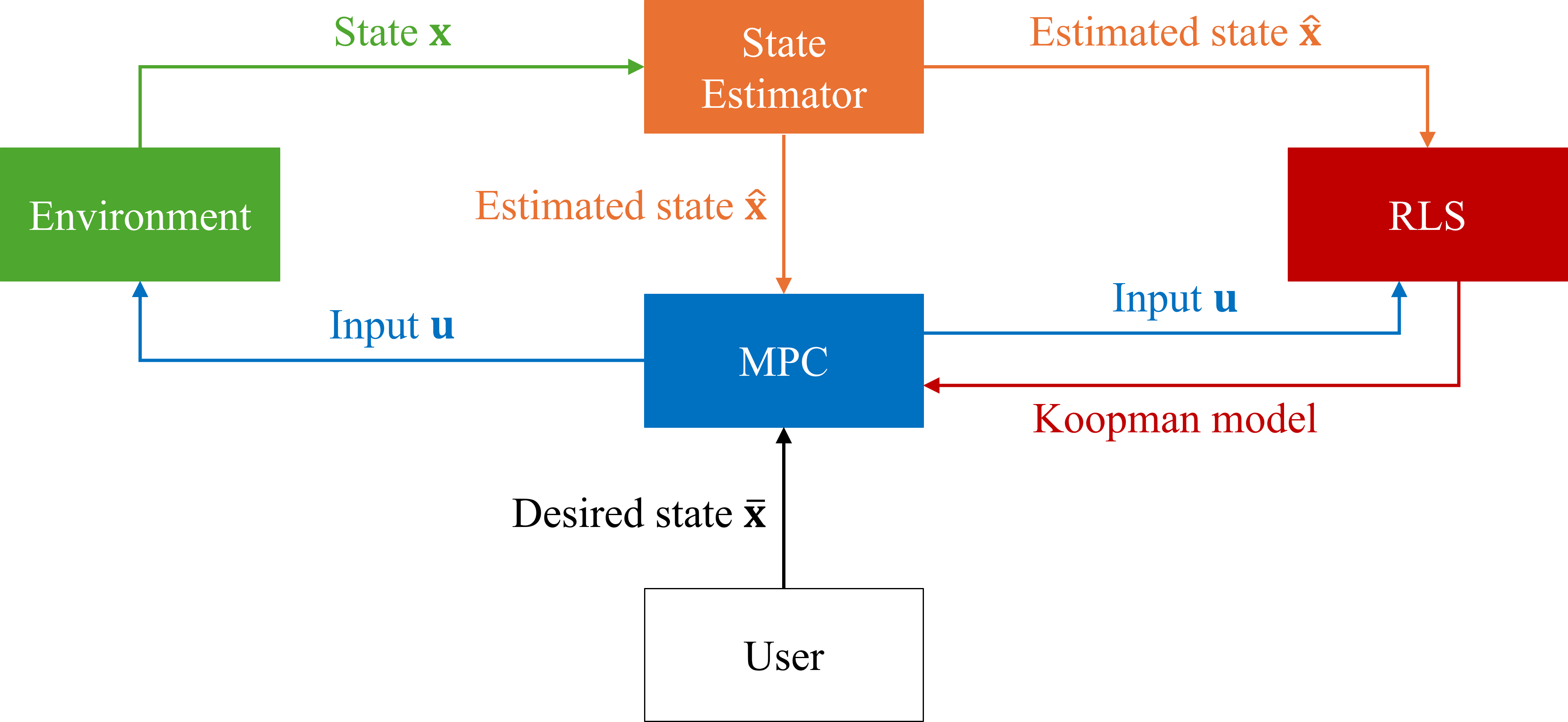}  
    \caption{Recursive Koopman Learning Pipeline}
    \label{fig:rkl_pipeline}
\end{wrapfigure}

Designing controllers for  nonlinear, dynamic robotic systems is challenging. Conventional model-based approaches rely on first principles to derive system models and then compute control inputs by solving constrained optimal control problems, often through trajectory optimization. While these methods are mathematically rigorous and have been successfully deployed on  robotic systems~\citep{kleff2021,grandia2023,meduri2023,le2024}, they exhibit inherent limitations: they are difficult to apply to systems that cannot be easily modeled using first principles, such as soft robots; they require significant time and effort to be adapted for highly dynamic tasks or complex environments with significant uncertainties; they struggle with designing controllers for hybrid dynamical systems, especially those involving multiple contact modes; and as the number of constraints increases, these methods impose escalating demands on computational resources and solver performance.

Data-driven approaches have advantages in dealing with uncertainties and complex dynamics. Over the past decade, these approaches---especially Reinforcement Learning (RL)---have made significant advances. RL algorithms~\citep{schulman2017,fujimoto2018,haarnoja2018} have been widely used to design controllers for various robotic systems and have achieved remarkable performance in challenging tasks, including soft robot control~\citep{jitosho2023}, dexterous manipulation~\citep{rajeswaran2018,chen2022}, vehicle control~\citep{williams2017}, and dynamic locomotion~\citep{joonho2020,mock2023}. However, RL typically require large amounts of data and have low sample efficiency, and their training demands substantial computational resources. These factors make it difficult to perform RL training solely on hardware---an important consideration for systems that cannot be simulated with high fidelity, such as soft robots. Additionally, due to the stochastic processes involved and the use of neural networks, RL often lack convergence guarantees and mathematical interpretability.

Given the disadvantages of conventional model-based control and RL, there is a need for sample-efficient and computationally lightweight data-driven control policy learning methods. A natural idea is to use linear model-based methods---linearity brings convexity and makes computations lighter and faster, and model-based methods tend to have higher sample efficiency. Standard linearization---which is performed locally and in the original state space---can only produce accurate local models. In contrast, Koopman theory can help us find \emph{exact linear models for nonlinear systems} in infinite dimensional spaces. With advances in computational capabilities, it has become feasible to estimate Koopman operators effectively using data-driven methods such as Dynamic Mode Decomposition (DMD) and Extended Dynamic Mode Decomposition (EDMD)~\citep{tu2013,williams2015}. As a result, Koopman theory has been successfully applied in various fields, including fluid dynamics~\citep{rowley2017}, physics~\citep{geneva2022}, power systems~\citep{susuki2010}, and robotics~\citep{bruder2019,bruder2024,Abaraham2019,mamakoukas2021,li2024}.

To improve sample efficiency, a key strategy is to identify and use the most informative data. A widely accepted but rarely formalized hypothesis---referred to here as the Attempting Control Goal (ACG) hypothesis---suggests that data collected while attempting a control objective with an ideal policy is particularly informative for that task. An ideal policy is a control policy that is dynamically feasible and globally optimizes a specified performance criterion---such as minimizing a cost function or maximizing reward---under the true system dynamics. Although such a policy is unattainable, researchers aim to approximate it and use the resulting demonstration data to train control policies. In imitation learning~\citep{pomerleau1988,hayes1996}, for example, human or animal instructors are treated as noisy surrogates of the ideal policy, and their successful demonstrations are used for training. 

However, certain systems are so challenging that even skilled humans struggle to provide consistently successful demonstrations, and the additional resources required to collect these data cannot be ignored. This motivates a weak ACG hypothesis: when a high-quality approximation of the ideal policy is unavailable, does the iterative loop of (i) computing the best policy attainable with the data so far and (ii) repeatedly attempting to control the system still yield data that are highly informative? This motivates us to adopt real-time online model updates.

Researchers have attempted updating NN models with online data~\citep{williams2017, nagabandi2020}, but these model-updating methods are heavy, slow, far from achieving real-time performance, and sample-inefficient. Linear models are particularly well-suited for this purpose, and researchers have explored updating Koopman models with new data. While a few papers mention model updates~\citep{Abaraham2019, li2024}, they primarily rely on retraining or iterative numerical solvers to update the parameters, which can be computationally intensive and slow. Additionally, some works have proposed faster online updating algorithms for linear models inspired by the Recursive Least Squares (RLS) algorithm~\citep{hayes1996, zhang2019, calderon2021}; however, these approaches lack comprehensive convergence and complexity analyses, and no experimental validation has been conducted on real-world robotic systems.~\citep{calderon2021} also uses RLS and MPC, but its update termination condition---which hinders convergence analysis---and MPC operate on the state space rather than the observable, questionable for Koopman-based methods.

Several studies have analyzed the convergence properties of EDMD. For instance, \citep{nuske2023,zhang2023,philipp2024} have provided detailed discussions on the convergence of EDMD for both stochastic and deterministic systems, under i.i.d. and ergodic sampling assumptions. However, no existing work formally analyzes the convergence of EDMD on Markov chains---a key property of data generated by dynamical systems. The original EDMD paper~\citep{williams2015} conceptually discusses approximations of the stochastic Koopman operator for Markov processes, but does not present a formal analysis. In addition, it omits the assumptions necessary to apply the Strong Law of Large Numbers for Markov chains~\citep{breiman1960}, and its treatment of the sampling density does not impose any assumptions on dataset structure.~\citep{nuske2023} derives probabilistic error bounds on finite data assuming \emph{i.i.d.} or ergodic sampling with the exponentially stable Koopman semigroup---a strong assumption which presupposes all non-constant observables decay exponentially toward their equilibrium values over time---without considering data correlation that is critical when using data from trajectories.

In this paper, we introduce Recursive Koopman Learning (RKL), a sample-efficient, Koopman-based learning pipeline. By leveraging Koopman theory, EDMD~\citep{williams2015}, RLS~\citep{hayes1996,zhang2019,calderon2021}, and Model Predictive Control (MPC), RKL enables rapid control policy learning and real-time recursive model updates. RKL significantly outperforms other data-driven methods in sample efficiency. On a simulated planar two-link arm, it surpasses RL benchmarks using $<$5\% of their data and with far lower computational cost. In hardware, RKL requires only 1 minute and 20 seconds (8,000 steps, 6,000 pretrained) to design a high-performance controller (Fig.~\ref{fig:box_plot}) for a highly nonlinear, hybrid-dynamic soft robot (Appx. Fig.~\ref{fig:soft_stewart_platform}). In contrast, a SotA RL algorithm takes 2 hours and 46 minutes (100,000 steps) and only attains $<$50\% of RKL's performance. We also provide the first formal convergence analysis of EDMD and RLS under continuous data growth within Markov chains, establishing sufficient conditions for convergence---crucial for practical use and previously unaddressed. We also discuss how to relate these conditions to the ACG hypothesis and our experimental results. Additionally, we present detailed algorithmic analyses of EDMD and RLS, demonstrating that RKL is computationally lightweight, with complexity independent of dataset size.

The contributions of the paper are summarized as follows:
\begin{itemize}
    \item We present RKL, an extremely sample-efficient, data-driven control policy learning pipeline with low computational cost, supported by rigorous convergence and algorithmic complexity analysis as well as thorough experiments on hardware and in simulation.
    \item We explicitly identify the sufficient conditions for the convergence of EDMD and RLS under continuous data growth in the context of Markov chains, addressing a gap in prior research. Additionally, we discuss the association between this analysis and the ACG hypothesis, and how they can guide the design of sample-efficient model learning methods.
    \item We provide a high-performance, multi-threaded C++ codebase designed to enable easy deployment and reproducibility: \href{https://github.com/zixinz990/recursive-koopman-learning.git}{https://github.com/zixinz990/recursive-koopman-learning.git}.
\end{itemize}

This paper is organized as follows: Sec.~\ref{sec:background} provides the necessary background. Sec.~\ref{sec:recursive_koopman_learning} details the RKL pipeline. Sec.~\ref{sec:experiments} presents our experimental results and discuss their relation to the ACG hypothesis and our convergence analysis. Sec.~\ref{sec:conclusion} summarizes our conclusions. Finally, Sec.~\ref{sec:limitations} discusses the limitations of our method and outlines directions for future research. The appendix is in the supplementary materials.

\section{Background}\label{sec:background}

This section briefly presents the mathematical background of the proposed method, including Koopman theory, EDMD, RLS and MPC.

\subsection{Koopman Theory}\label{sec:koopman_theory}

Consider a discrete-time dynamical system $\mathbf{x}_{k+1}=\mathbf{f}\left(\mathbf{x}_k\right)$, $\mathbf{x}\in\mathbf{X}\subset\mathbb{R}^{n_x}$. Define an observation function (or ``observable'') $\mathbf{z}:=\boldsymbol{\phi}\left(\mathbf{x}\right): \mathbb{R}^{n_x}\rightarrow\mathbb{R}^{n_z},n_z\geq n_x$, which lifts the state into a higher-dimensional space. Let $\mathbf{H}$ be a Hilbert space on $\mathbf{X}$. If the components of the observation function $\left\{\boldsymbol{\phi}_1,\boldsymbol{\phi}_2,...\right\}$ form an orthonormal set of basis functions spanning $\mathbf{H}$, and the composition function $\boldsymbol{\phi}\circ\mathbf{f}$ is also an element of the Hilbert space, then there exists a linear operator $\mathbf{K}$ such that $\boldsymbol{\phi}\circ\mathbf{f}\left(\mathbf{x}\right)=\mathbf{K}\boldsymbol{\phi}\left(\mathbf{x}\right)$, when $n_z=\infty$~\citep{koopman1931,asada2023}.
We also need consider control inputs $\mathbf{u}\in\mathbb{R}^{n_u}$ to design controllers. Define another observation function $\mathbf{g}:=\boldsymbol{\psi}\left(\mathbf{u}\right):\mathbb{R}^{n_u}\rightarrow\mathbb{R}^{n_g},n_g\geq n_u$ and a new pair of variables $\boldsymbol{\alpha}_k\in\mathbb{R}^{n_z+n_g}=\left[\boldsymbol{\phi}\left(\mathbf{x}_k\right),\boldsymbol{\psi}\left(\mathbf{u}_k\right)\right]$ and $\boldsymbol{\beta}_k\in\mathbb{R}^{n_z+n_g}=\left[\boldsymbol{\phi}\left(\mathbf{x}_{k+1}\right),\boldsymbol{\psi}\left(\mathbf{u}_k\right)\right]$, 
the Koopman model becomes $\boldsymbol{\beta}_k=\mathbf{K}\boldsymbol{\alpha}_k$~\citep{Abaraham2019,proctor2016}. We can extract $\mathbf{K}_z$ and $\mathbf{K}_g$ from $\mathbf{K}$ and get
\begin{equation}\label{eq:koopman_w_control}
    \boldsymbol{\phi}\left(\mathbf{x}_{k+1}\right)=\mathbf{K}_z\boldsymbol{\phi}\left(\mathbf{x}_k\right)+\mathbf{K}_g\boldsymbol{\psi}\left(\mathbf{u}_k\right).
\end{equation}

\subsection{Extended Dynamic Mode Decomposition}\label{sec:edmd}

We need to estimate the Koopman operator using data in a finite-dimensional observation space in order to use it. First, define two data matrices $\mathbf{Y}=\left[\boldsymbol{\alpha}_0~\boldsymbol{\alpha}_1~\cdots\right]\in\mathbb{R}^{\left(n_z+n_g\right)\times N}$ and $\bar{\mathbf{Y}}\left[\boldsymbol{\beta}_0~\boldsymbol{\beta}_1~\cdots\right]\in\mathbb{R}^{\left(n_z+n_g\right)\times N}$, where $N$ is the number of data. $\mathbf{Y}$ contains the ``current'' data, and $\bar{\mathbf{Y}}$ contains the data one time step further. We can estimate the Koopman matrix $\mathbf{K}$ by solving
\begin{equation}\label{eq:edmd_optimization}
    \min_\mathbf{K}||\mathbf{K}\mathbf{Y}-\bar{\mathbf{Y}}||_F,
\end{equation}
where $F$ indicates the Frobenius norm. To solve this, EDMD does
\begin{equation}\label{eq:edmd}
    \mathbf{K}=\bar{\mathbf{Y}}\mathbf{Y}^\top\left(\mathbf{Y}^\top\right)^\dagger\mathbf{Y}^\dagger=\left(\bar{\mathbf{Y}}\mathbf{Y}^\top\right)\left(\mathbf{Y}\mathbf{Y}^\top\right)^\dagger.
\end{equation}
The matrix $\mathbf{Y}\mathbf{Y}^\top$ is a square matrix and is well-conditioned to matrix inversion computation.

\subsection{Recursive Least Squares}\label{sec:rls}

Suppose we have a standard linear regression model of the form $y_k=\mathbf{x}_k^\top\mathbf{w}+e_k$, where $y\in\mathbb{R}$ is the scalar output, $\mathbf{x}\in\mathbb{R}^n$ is the input/regressor vector, $\mathbf{w}\in\mathbb{R}^n$ is the unknown parameter vector we wish to estimate, $e\in\mathbb{R}$ is noise or modeling error. To estimate $\mathbf{w}$ in a way that incorporates each new data pair $\left(\mathbf{x}_k, y_k\right)$, RLS algorithm solves $\min_\mathbf{w}J=\sum_{k=1}^N\left(y_k-\mathbf{x}_k^\top\mathbf{w}\right)$. First, initialize the RLS algorithm at $k=0$ by picking $\mathbf{w}_0$ and $\mathbf{P}_0$. At each time step $k$, given the new data pair $\left(\mathbf{x}_k, y_k\right)$, update the gain vector $\mathbf{g}_k$, the parameter estimate $\mathbf{w}_k$, and the inverse covariance $\mathbf{P}_k$ by
\begin{equation}\label{eq:rls_update}
    \mathbf{g}_k=\frac{\mathbf{P}_{k-1}\mathbf{x}_k}{1+\mathbf{x}_k^\top\mathbf{P}_{k-1}\mathbf{x}_k}, \mathbf{w}_k=\mathbf{w}_{k-1}+\mathbf{g}_k\left(\mathbf{y}_k-\mathbf{x}_k^\top\mathbf{w}_{k-1}\right), \mathbf{P}_k=\mathbf{P}_{k-1}-\frac{\mathbf{P}_{k-1}\mathbf{x}_k\mathbf{x}_k^\top\mathbf{P}_{k-1}}{1+\mathbf{x}_k^\top\mathbf{P}_{k-1}\mathbf{x}_k}.
\end{equation}
Conceptually, $\mathbf{P}_k$ tracks the inverse of the ``data covariance'' seen so far. Updating $\mathbf{P}$ uses the Sherman-Morrison formula~\citep{sherman1950} so that large matrix inversions are unnecesary. In Sec.~\ref{sec:rt_recursive_model_updates} and Appx.~\ref{sec:rls=edmd}, we show that RLS can be generalized to update state-space models.

\subsection{Model Predictive Control}\label{sec:mpc}


A classical MPC policy computes control actions by solving a trajectory optimization problem at each step. It minimizes an objective function composed of a terminal cost and either an integral (for continuous-time) or a summation (for discrete-time) of the running cost, subject to initial conditions, system dynamics, and equality and inequality constraints. Only the first control input from the solution is applied before the process repeats at the next time step. More details are provided in Appx.~\ref{sec:mpc-sac}.

\section{Recursive Koopman Learning}\label{sec:recursive_koopman_learning}

\newlength{\intextsepdefault} 
\setlength{\intextsepdefault}{\intextsep} 
\setlength{\intextsep}{0pt}
\begin{wrapfigure}[12]{r}{0.518\textwidth}
    \centering
    \begin{minipage}{0.518\textwidth}
    \begin{algorithm}[H]
    \caption{Recursive Koopman learning}\label{alg:rkl}
        \begin{algorithmic}
            \Require Initial dataset $\mathbf{Y}_0$ and $\mathbf{Y}_1$
            \State $\mathbf{K}_0,\mathbf{P}_0\gets\text{EDMD}\left(\mathbf{Y}_0, \mathbf{Y}_1\right)$
            \State $\mathbf{u}_0=\text{MPC}\left(\mathbf{K}_0,\mathbf{z}_0,\bar{\mathbf{z}}\right)$\Comment{$\bar{\mathbf{z}}$ is the goal}
            \State $\mathbf{z}_1=\text{ENV}\left(\mathbf{z}_0,\mathbf{u}_0\right)$
            \State $k=0$
            \While{True}
                \State $\mathbf{K}_{k+1},\mathbf{P}_{k+1}\gets\text{RLS}\left(\mathbf{K}_k,\mathbf{P}_k,\mathbf{z}_k,\mathbf{u}_k,\mathbf{z}_{k+1}\right)$
                \State $\mathbf{u}_{k+1}\gets\text{MPC}\left(\mathbf{K}_{k+1},\mathbf{z}_{k+1},\bar{\mathbf{z}}\right)$
                \State $\mathbf{z}_{k+2}=\text{ENV}\left(\mathbf{z}_{k+1},\mathbf{u}_{k+1}\right)$
                \State $k\gets k+1$
            \EndWhile
        \end{algorithmic}
    \end{algorithm}
    \end{minipage}
\end{wrapfigure}

In this section, we present the details of RKL, which is built on an MPC framework. Initially, a small dataset is collected offline and used to compute an initial model via EDMD, which serves to initialize both the MPC and RLS modules. Subsequently, the MPC controller, using the Koopman model, iteratively computes the optimal control input at each time step according to the user-defined control objective. Simultaneously, the RLS module continuously updates the Koopman model in real time using newly collected data. The overall pipeline is shown in Fig.~\ref{fig:rkl_pipeline} and Alg.~\ref{alg:rkl}.

\subsection{MPC Solver}\label{sec:mpc_solver}

For controller synthesis, we formulate an MPC problem using the learned Koopman model. Any MPC technique could be used; we use Sequential Action Control (MPC-SAC)~\citep{Ansari2016} as the solver because it achieves better performance than LQR in experiments (details in Sec.~\ref{sec:experiments}), similar to the findings in~\citep{nishimura2021}. Further details can be found in~\citep{Ansari2016} and Appx.~\ref{sec:mpc-sac}.

\subsection{Dataset Initialization}\label{sec:dataset initialization}

Previous studies have initialized Koopman models using normal distributions~\cite{Abaraham2019}, but this often causes numerical issues, such as severely ill-conditioned optimal control problems. In MPC-SAC, such an initialization can cause the adjoint variable to become excessively large, leading to overflow. To avoid this, we initialize the Koopman model using a dataset, ensuring numerical stability during MPC optimization. Sec.~\ref{sec:experiments} details the initial dataset collection and its impact on control performance.

\subsection{Real-Time Recursive Model Update}\label{sec:rt_recursive_model_updates}

The learned Koopman model is continuously updated while pursuing the control goal. To efficiently utilize runtime data, we aim to find a stable, fast, and lightweight real-time model updating algorithm. RLS (introduced in Sec.~\ref{sec:rls}) is well-suited for this purpose: it is tailored for linear systems, much faster and lighter than resolving full linear regressions, and can be readily generalized to update linear state-space models. Appx. Alg.~\ref{alg:rls} summarizes the algorithm.

Consider the two data matrices used in Eq.~\ref{eq:edmd_optimization}. At time step $k$, we have the data matrices $\mathbf{Y}_k$ and $\bar{\mathbf{Y}}_k$, as well as the latest estimate of the Koopman matrix $\mathbf{K}_k$. Additionally, we define
\begin{equation}\label{eq:def_Qk_Pk}
    \mathbf{Q}_k:=\bar{\mathbf{Y}}_k\mathbf{Y}_k^\top, \mathbf{P}_k:=\left(\mathbf{Y}_k\mathbf{Y}_k^\top\right)^\dagger.
\end{equation}
After one time step, we obtain a new pair of data ${\boldsymbol{\alpha}_{k}, \boldsymbol{\beta}_{k}}$ and we use them to update the model $\mathbf{K}_k$. Finding that $\mathbf{P}$ tracks the inverse of the data covariance, similar to RLS (Eq.~\ref{eq:rls_update}), we can utilize the Sherman-Morrison formula~\citep{sherman1950} and the update rule of $\mathbf{P}$ is given by
\begin{equation}\label{eq:update_Pk_gammak}
    \mathbf{P}_{k+1}=\mathbf{P}_k-\gamma_k\mathbf{P}_k\boldsymbol{\alpha}_k\boldsymbol{\alpha}_k^\top\mathbf{P}_k, \gamma_k:=\frac{1}{1+\boldsymbol{\alpha}_k^\top\mathbf{P}_k\boldsymbol{\alpha}_k}.
\end{equation}
It can be shown that the estimate of the Koopman matrix can be updated through
\begin{equation}
    \mathbf{K}_{k+1}=\mathbf{K}_k+\gamma_k\left(\boldsymbol{\beta}_k-\mathbf{K}_k\boldsymbol{\alpha}_k\right)\boldsymbol{\alpha}_k^\top\mathbf{P}_k.
\end{equation}

A single RLS update yields \emph{exactly the same result as retraining the entire model with EDMD} on the updated dataset (see Appx.~\ref{sec:rls=edmd} for the proof). Our algorithmic analysis (Appx.~\ref{sec:algorithmic_complexity_analysis}) shows that RLS is well-suited for online updates under continuous data growth, as \emph{its speed depends only on the dimension of the observables}, not on dataset size. Additionally, among all methods that compute exact solutions to the linear regression in Eq.~\ref{eq:edmd_optimization}, RLS has the lowest time complexity. In contrast, EDMD’s complexity grows with both the observables dimension and dataset size, making it increasingly slower as more data is added, which is detrimental to real-time model updates under continuous data growth.

Based on the conclusion that RLS and EDMD produce identical computational results, we discuss the convergence properties of RKL by analyzing the convergence of EDMD. In particular, we assume that the dataset forms a Markov chain, which aligns well with our data collection process (Sec.~\ref{sec:experiments}) for both the initial and online dataset.

We find that the Strong Law of Large Numbers for Markov chains~\citep{breiman1960} can be used to prove the convergence of EDMD under continuous data growth. However, this theorem requires the dataset to be an \emph{irreducible}, \emph{aperiodic}, and \emph{positively recurrent} Markov chain w.r.t. an invariant measure $\mu$, meaning the chain is \emph{ergodic}. These requirements are overlooked in previous research~\citep{williams2015}. In addition, the observation function must be $\mu$-square-integrable, and the $\mathbf{P}$ matrix in Eq.~\ref{eq:def_Qk_Pk} must be of full rank. Together, these conditions provide sufficient guarantees for the convergence of EDMD as the size of the dataset approaches infinity. Details of the analysis is presented in Appx.~\ref{sec:convergence_analysis}. In the next section, we relate the above results to the experimental results and the ACG hypothesis.

\section{Experiments}\label{sec:experiments}

\setlength{\intextsep}{0pt}
\begin{wrapfigure}[32]{r}{0.4\textwidth}
    \centering
    \includegraphics[width=\linewidth]{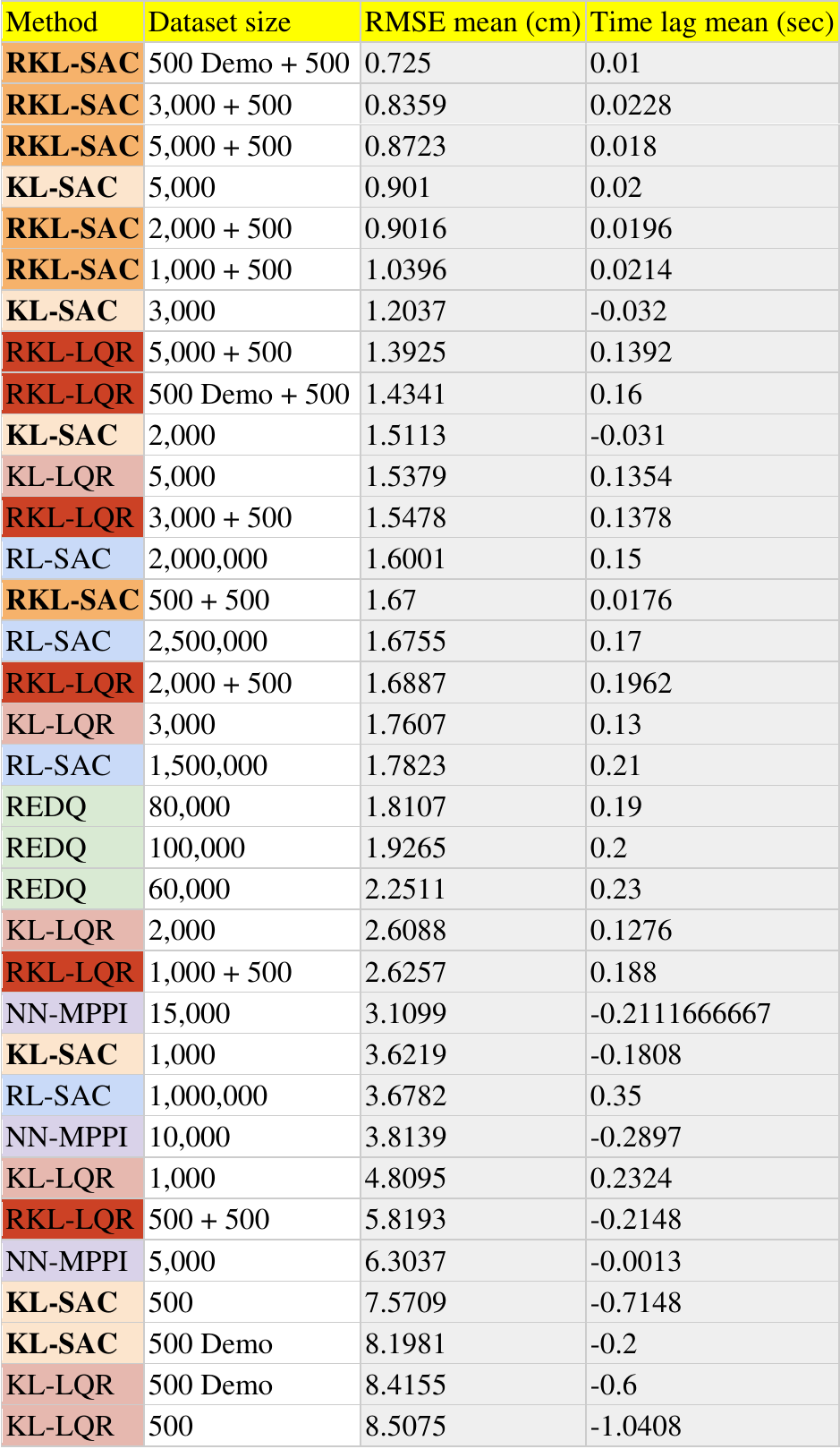}
    \vspace{-0.2cm}
    \captionsetup{type=table}
    \caption{The results of tracking a reference trajectory on the simulated planar two-link arm, sorted by RMSE. Online model updates significantly improves sampling efficiency and control performance. Additionally, MPC-SAC enhances the performance of both RKL and KL compared to using LQR. RKL shows higher sample efficiency and superior control performance than the baselines.}
    \label{fig:reacher_data_table}
\end{wrapfigure}

Our experiments are conducted on a simulated tracking tasks with a planar two-link arm, and dynamic balancing with a Soft Stewart Platform (SSP)~\citep{ketchum2025} (Appx. Fig.~\ref{fig:soft_stewart_platform}), a nonlinear hybrid system actuated with soft transmissions~\citep{kim2024}. Details are shown in Appx.~\ref{sec:details_of_experiments}.

\subsection{Benchmarks}

We use the following five methods in our experiments:
\begin{itemize}
    \item \textbf{RKL-SAC} and \textbf{RKL-LQR}: Both are RKL, but different MPC solvers are chosen (MPC-SAC or LQR).
    \item \textbf{KL-SAC} and \textbf{KL-LQR}: Standard Koopman learning without any model updates~\citep{Abraham2017}.
    \item \textbf{NN-MPPI}: Model-based RL using Model Predictive Path Integral (MPPI) control via NN models~\citep{williams2017}.
    \item \textbf{RL-SAC}: Soft Actor-Critic, a prominent model-free RL algorithm~\citep{haarnoja2018}.
    \item \textbf{REDQ}: Randomized Ensembled Double Q-Learning, a SotA sample-efficient model-free RL algorithm~\citep{chen2021}.
\end{itemize}
\vspace{-10pt}


\subsection{Planar Two-Link Arm Experiments}\label{sec:planar_two_link_arm_experiments}

We evaluate all benchmarks on a planar two-link arm simulated in MuJoCo. The goal is to control the joint states to any desired states in the workspace, and the performance is assessed based on the ability to control the tip to track a 500-step reference trajectory shaped like an ``8'' (details are shown in Appx.~\ref{sec:planar_two_link_arm} and Fig.~\ref{fig:kl_rkl_8_traj}-\ref{fig:redq_8_traj}). Each method is tested across 50 trials, summarized in Table~\ref{fig:reacher_data_table}. RKL achieves the best overall performance while requiring a significantly smaller dataset compared to RL. Specifically, RKL-SAC outperforms the RL baselines using only 3,500 steps of data---3,000 of which are generated by random actions---representing just $\sim$5\% of the minimum data required by RL methods. Although NN-MPPI requires less data than model-free RL methods (though still more than RKL), its control performance is comparatively lower. Furthermore, using MPC-SAC improves the performance of both RKL and KL relative to LQR, supporting our claim in Sec.~\ref{sec:mpc_solver}. Comparing KL to RKL demonstrates the impact of online recursive model updates---reducing not only the RMSE by up to 77.94\% but also the time lag, especially on small datasets.

We evaluate two types of initial datasets for KL and RKL: a 500-step demonstration dataset (500 Demo) and datasets generated by uniformly sampling actions from the action space. When initialized with the 500 Demo dataset, RKL-SAC outperforms all competing methods, including RKL-SAC trained on larger datasets. This advantage reflects the ACG hypothesis and the sufficient conditions for convergence discussed in Sec.~\ref{sec:rt_recursive_model_updates} and Appx.~\ref{sec:convergence_analysis}. In our convergence analysis, the state can be interpreted broadly to include both the system state and the control input, so that the controller can be treated as part of the dynamics. A closed-loop dynamical system with an ideal policy in the ACG hypothesis for a given control objective can be represented as an autonomous dynamical system in this generalized state space---\emph{the exact target system we aim to learn}. A high-quality demonstration induces a Markov chain that is nearly ergodic w.r.t. the target dynamics' invariant measure. As a result, the initial model’s invariant measure closely approximates the true one, providing a good start for rapid convergence to the optimal representation through online model updates.

The success of RKL empirically validates the weak ACG hypothesis, and our convergence analysis helps explain this. While pursuing the control objective, real-time online model updates ensure that the model estimate at each time step is globally optimal, in the sense of Eq.~\ref{eq:edmd_optimization}, given the currently available data. This also implies that the system is approaching the behavior of a perfect system in a manner we define as optimal at the moment---that is, it is doing its best to converge toward the ergodicity condition (though this is never attainable). We infer that metrics related to ergodicity should be developed to guide the data collection process, thereby ensuring it more closely aligns with the ergodicity requirements. This, in turn, would lead to faster convergence and more sample-efficient learning, particularly for tasks where collecting high-quality demonstration data is challenging.


\subsection{Soft Stewart Platform Experiments}\label{sec:ssp_experiments}

Due to the nature of the soft actuators and the mechanical design of the system, the SSP is a highly nonlinear, hybrid, and time-varying dynamical system, making it an ideal platform for testing RKL. A puck---comprised of a solid 38‑mm Delrin ball enclosed within a 5‑cm‑diameter plastic ring---is placed on the SSP. The objective is to control the puck’s state, enabling it to balance at a desired location or follow a reference trajectory. Details is presented in Appx.~\ref{sec:soft_stewart_platform}.

The initial dataset is collected by a human operator using a SpaceMouse Compact. The same initial dataset is used for all experiments but cropped to different time lengths (Appx.~\ref{sec:initial_dataset}). In RKL, each RLS update is completed in 20 ms (limited by the feedback thread).

\subsubsection{Puck Balancing}\label{sec:puck_balancing}

\setlength{\intextsep}{\intextsepdefault}
\begin{figure}[htbp]
    \centering
    \begin{subfigure}[b]{0.49\linewidth}
        \centering
        \includegraphics[width=\linewidth]{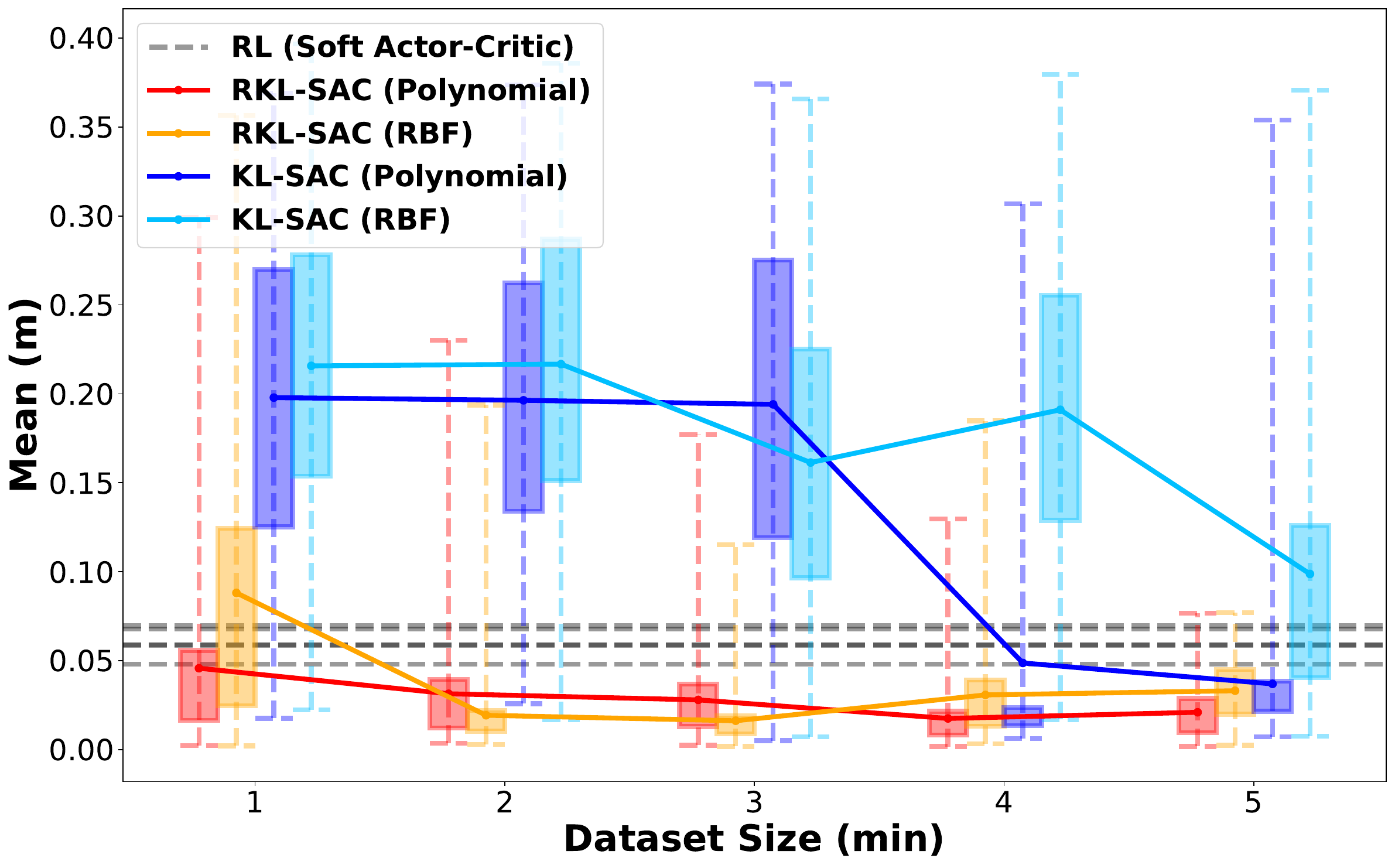}
        \caption{Mean}
        \label{fig:mean_abs_err}
    \end{subfigure}
    \hfill
    \begin{subfigure}[b]{0.49\linewidth}
        \centering
        \includegraphics[width=\linewidth]{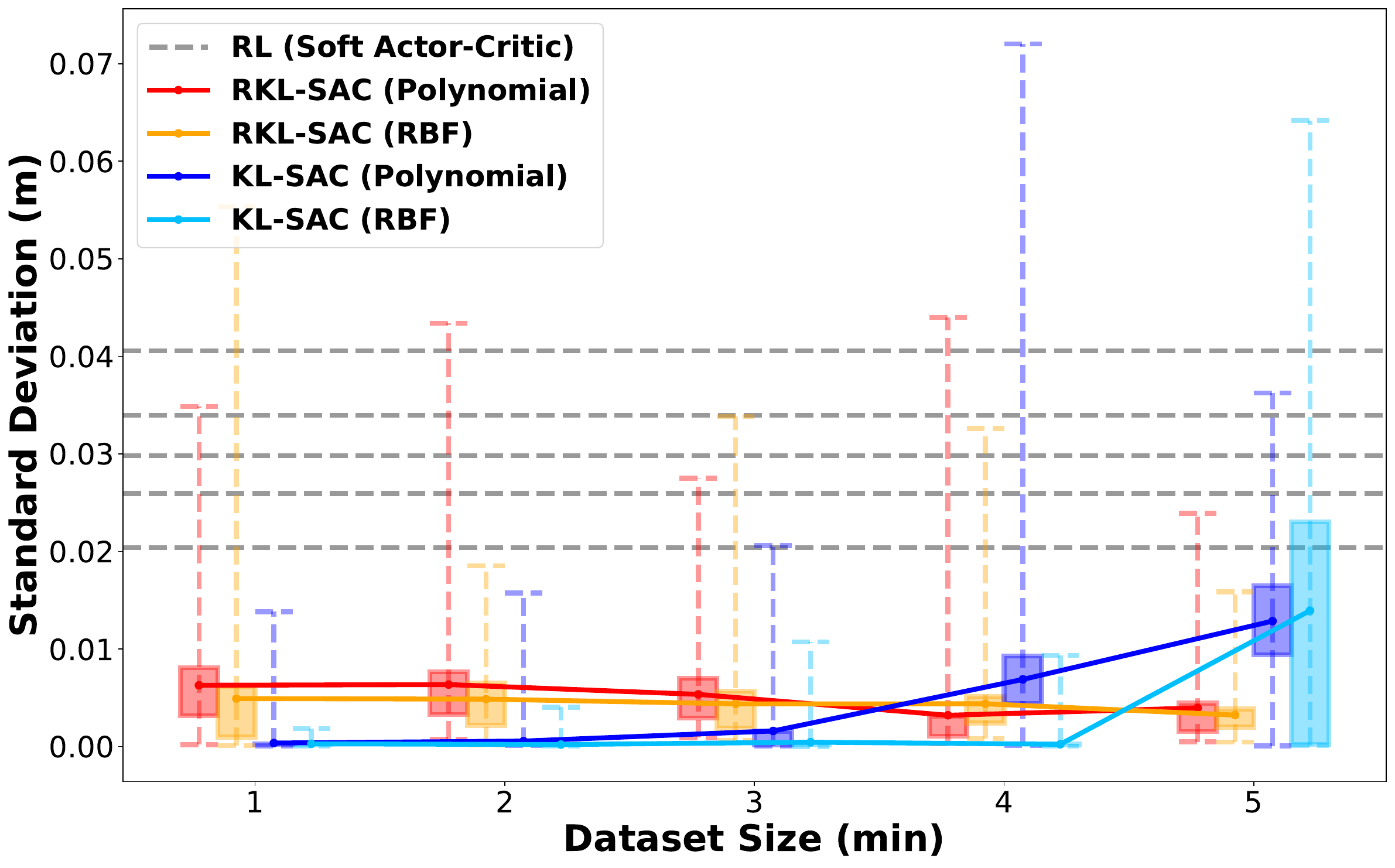}
        \caption{Standard Deviation}
        \label{fig:std_dev_abs_err}
    \end{subfigure}
    \caption{KL-SAC vs. RKL-SAC vs. RL-SAC (five random seeds) in puck balancing. The whiskers show the range of the data, the box represents the middle 50\%, and the lines connect the averages.}
    \label{fig:box_plot}
\end{figure}

\begin{wrapfigure}[16]{r}{0.5\textwidth}
    \centering
    \begin{subfigure}[b]{0.265\textwidth}
        \centering
        \includegraphics[width=\linewidth]{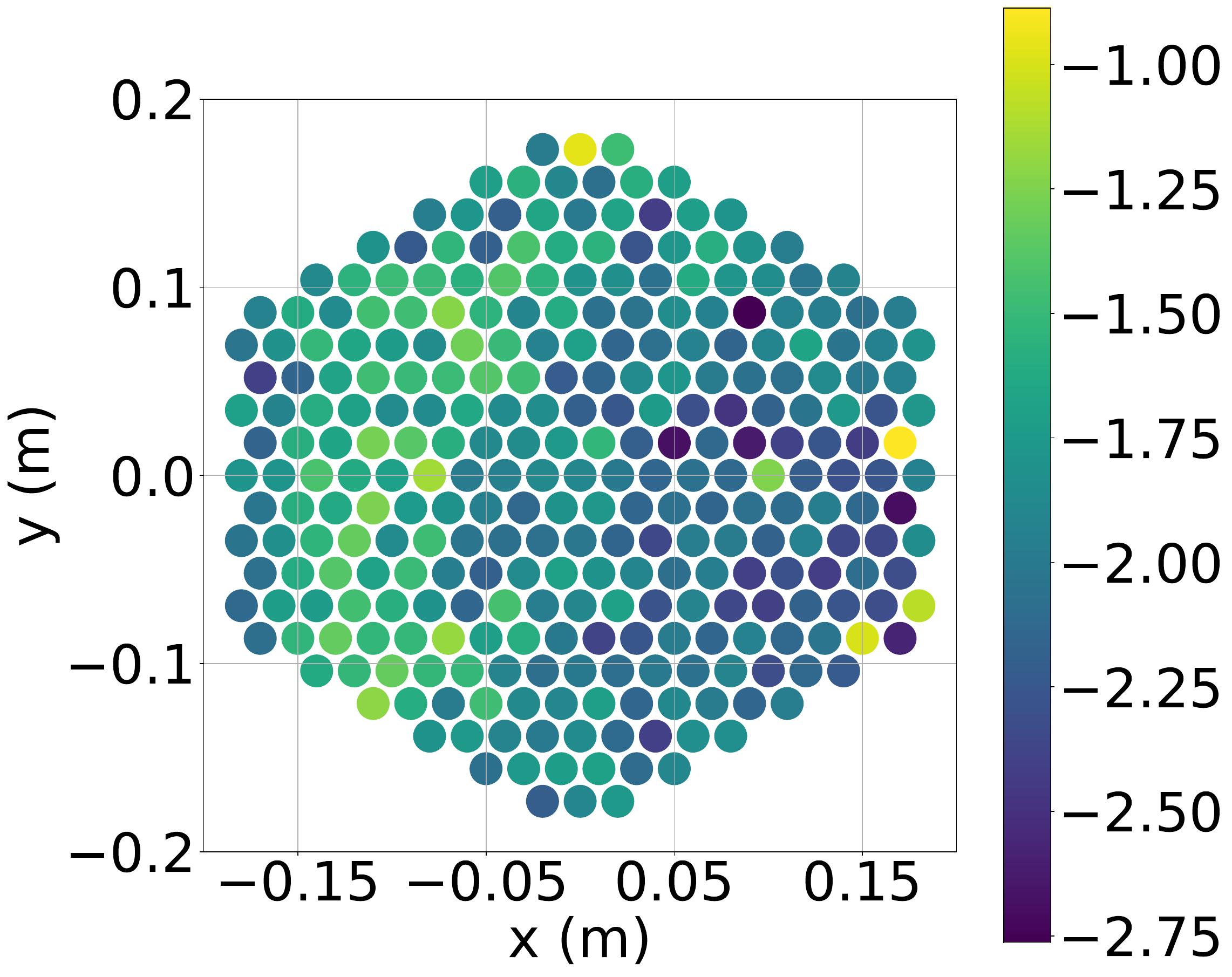}
        \caption{Mean}
        \label{fig:mean_abs_err_1_trial}
    \end{subfigure}
    \begin{subfigure}[b]{0.215\textwidth}
        \centering
        \raisebox{0.0cm}{\includegraphics[width=\linewidth]{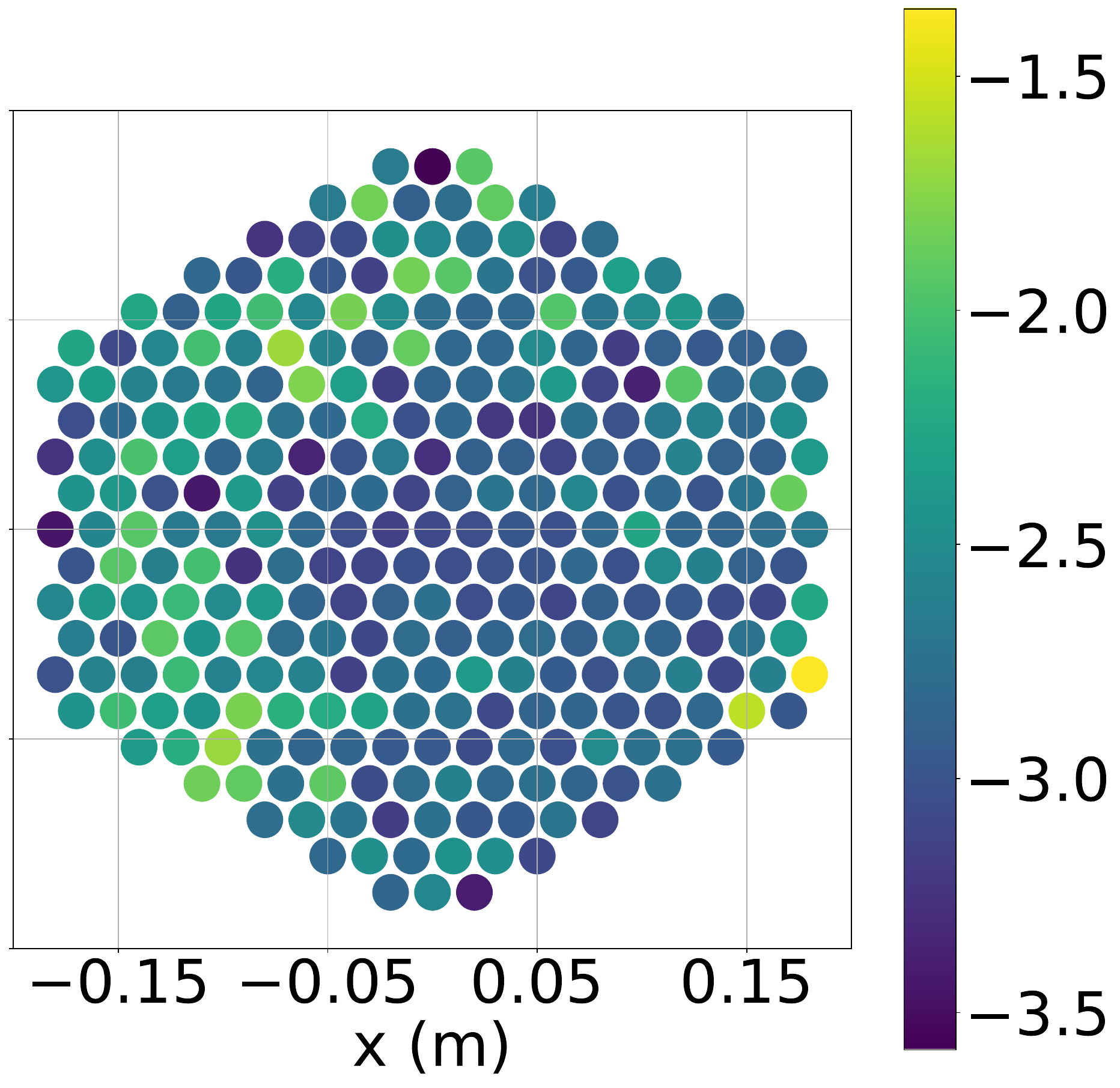}}
        \caption{Standard Deviation}
        \label{fig:std_dev_abs_err_1_trial}
    \end{subfigure}
    \caption{The mean and standard deviation of the absolute errors (scaled by log base 10) computed over the last 5 seconds of each 20-second RKL-SAC trial. This controller uses a 4-minute-long initial dataset and a 28-dimensional polynomial observation function. The distance between each goal point and its six adjacent goal points is 2 cm.}
    \label{fig:rkl_balance_ball_distribution}
\end{wrapfigure}

We test the learned controllers at 293 uniformly distributed goal points across the platform surface. Each trial lasts 20 seconds, and the model in RKL is reset at the end of each trial. The mean and standard deviation of the last-5-second absolute position error are used to evaluate the performance at each desired position. Large mean values indicate significant steady-state errors, while large standard deviations reflect substantial oscillations. Two types of observation functions are tested: a 28-dimensional polynomial basis function up to the third degree and a 117-dimensional Gaussian radial basis function (RBF) (Appx.~\ref{sec:kl_rkl_observation_functions}). Fig.~\ref{fig:rkl_balance_ball_distribution} illustrates a successful example of RKL-SAC.

We compare KL-SAC, RKL-SAC, and RL-SAC across varying dataset sizes. As shown in Fig.~\ref{fig:mean_abs_err}, KL-SAC struggles with datasets under 4 minutes, while RKL-SAC remains robust except for using RBF with 1-minute data. Fig.~\ref{fig:std_dev_abs_err} shows KL-SAC and RKL-SAC have low standard deviations, indicating stable puck behavior, whereas RL-SAC results in larger oscillations. However, KL-SAC’s low deviation with small datasets reflects puck immobilization due to static friction, not true stability---online updates help overcome this. Notably, RKL-SAC achieves better performance than RL-SAC using just 1 minute of initial data and 20 seconds of online updates (8,000 steps), compared to RL-SAC’s 2 hours 46 minutes (100,000 steps)---only 0.8\% of the training time. We also test NN-MPPI under the same setup~\citep{avtges2025}. RKL improves upon NN-MPPI with a 66\% lower average error and a 8 times better standard deviation, using only 2.6\% of the training time.

\setlength{\intextsep}{0pt}

\begin{figure}[htbp]
    \centering
    \begin{subfigure}[b]{0.215\linewidth}
        \centering
        \includegraphics[width=\linewidth]{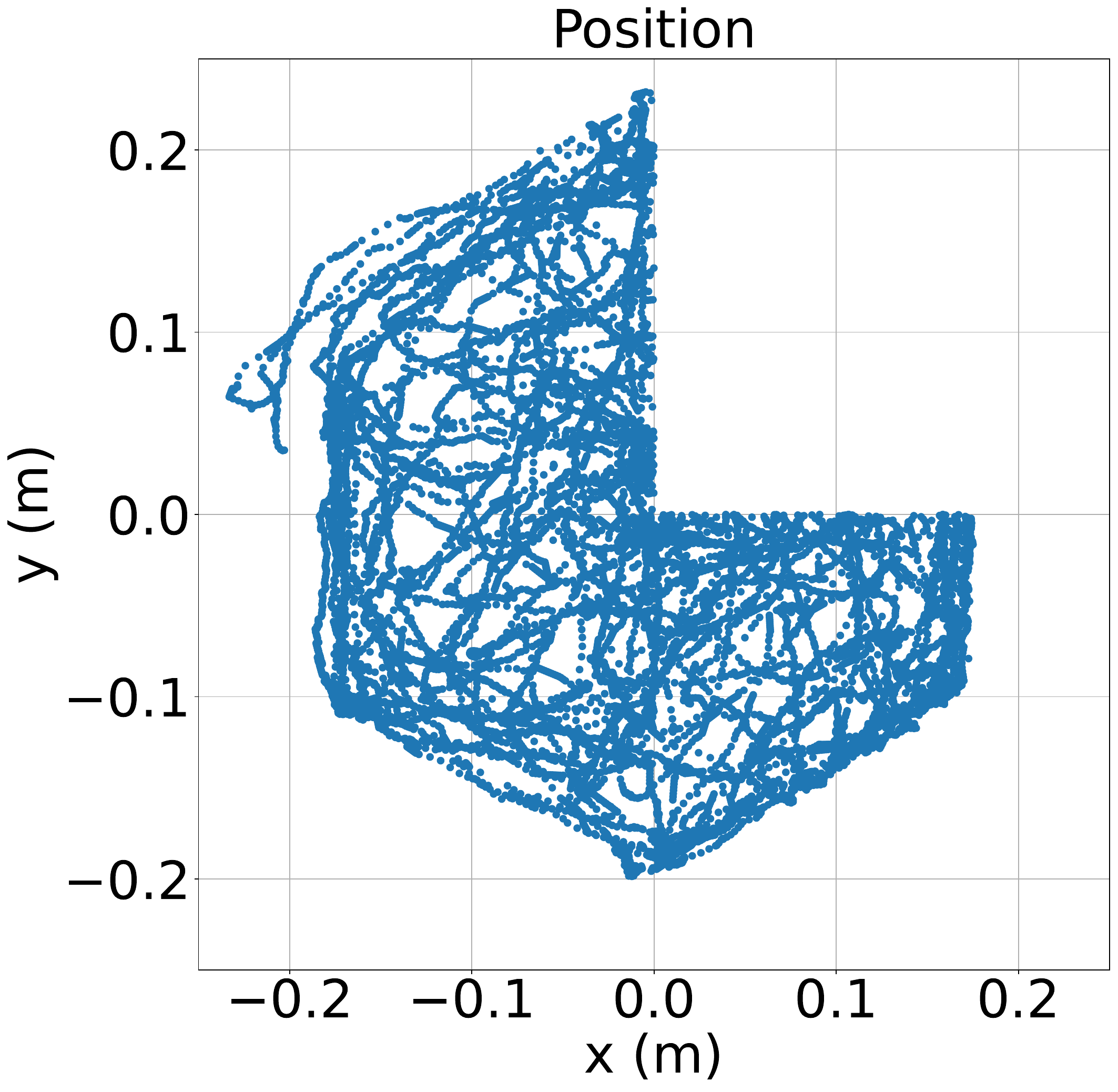}
        \caption{Position}
        \label{fig:cropped_dataset_position}
    \end{subfigure}
    \begin{subfigure}[b]{0.21\linewidth}
        \centering
        \includegraphics[width=\linewidth]{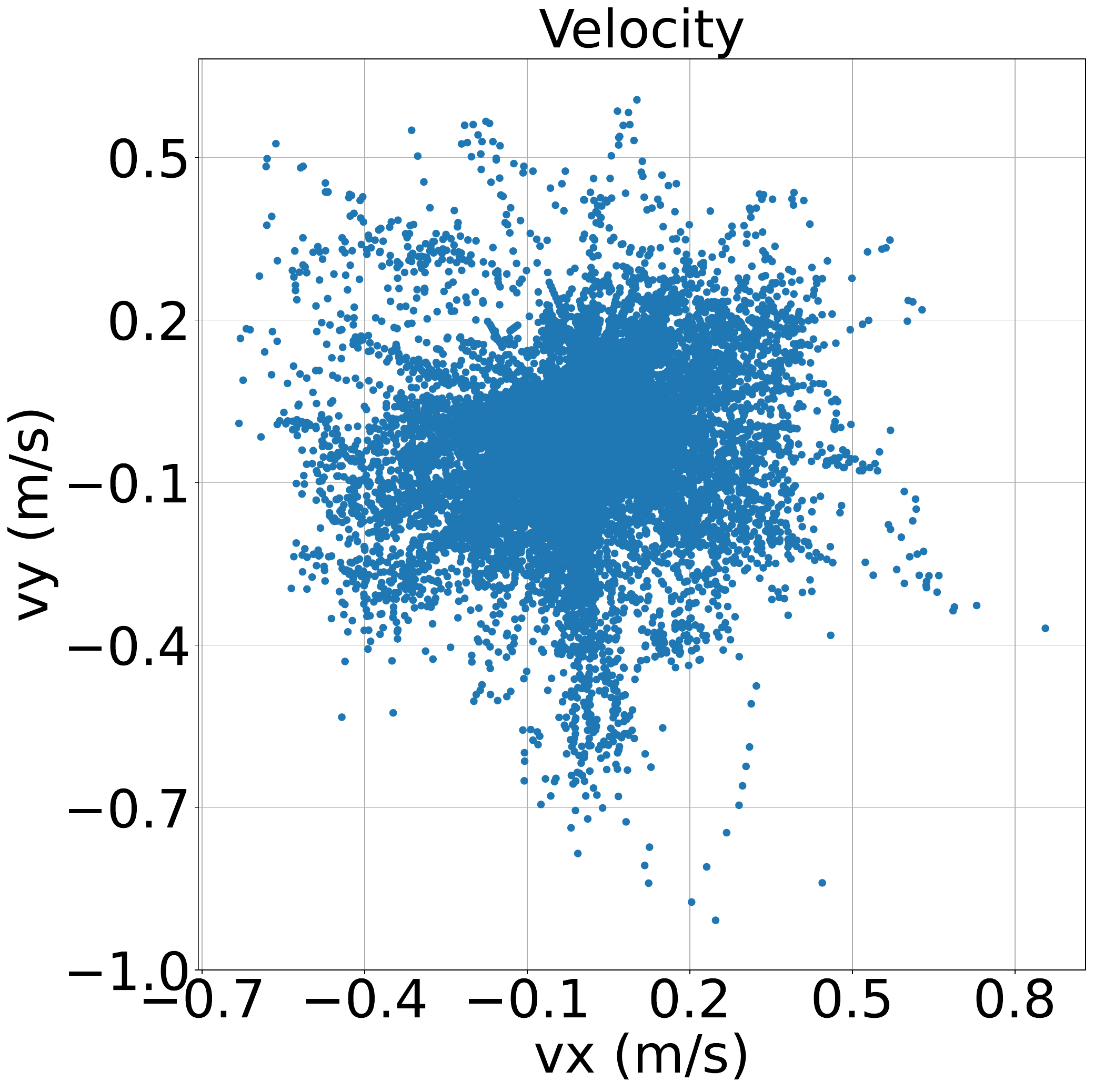}
        \caption{Velocity}
        \label{fig:cropped_dataset_velocity}
    \end{subfigure}
    \begin{subfigure}[b]{0.275\linewidth}
        \centering
        \includegraphics[width=\linewidth]{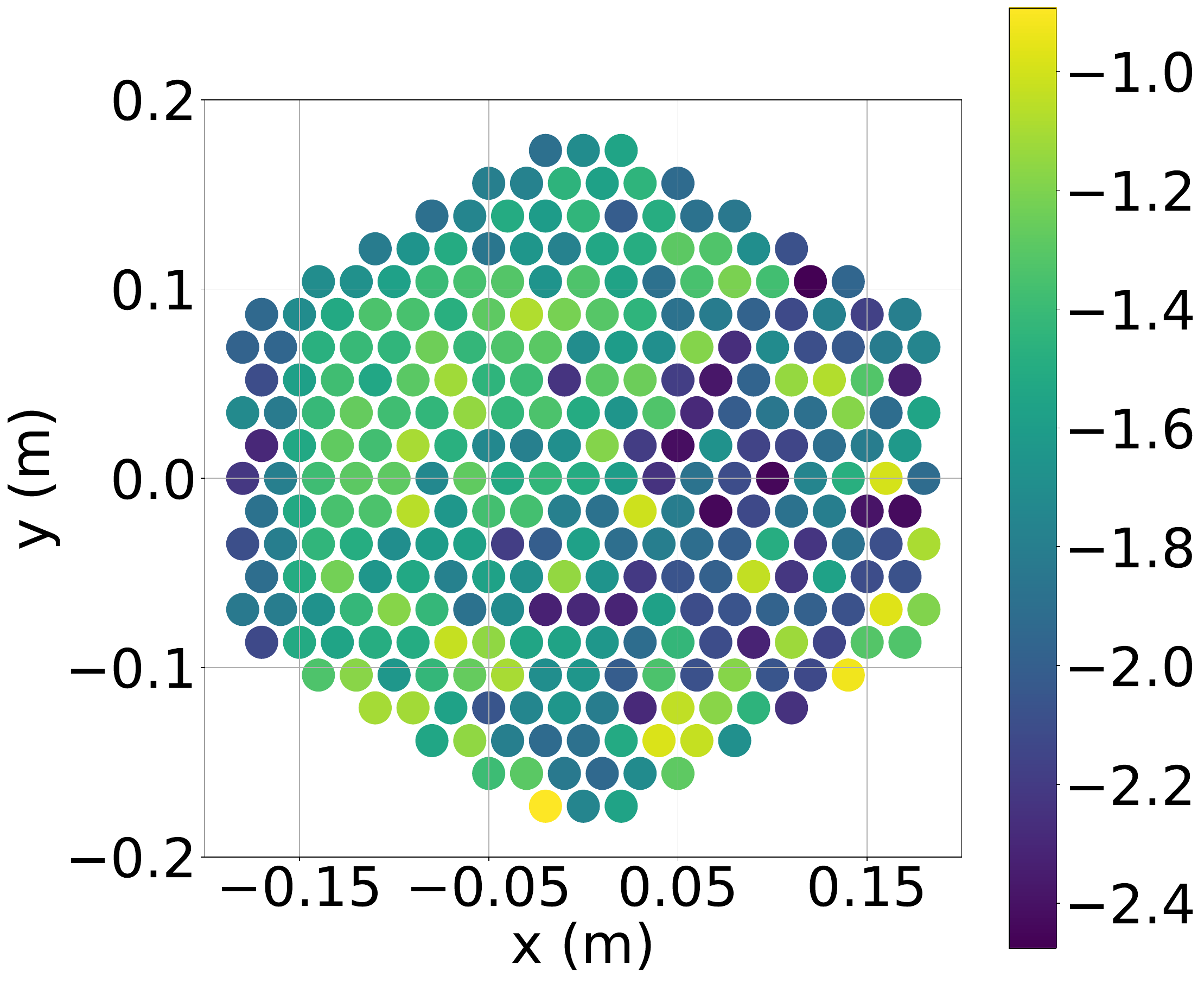}
        \caption{Mean}
        \label{fig:mean_abs_err_cropped_dataset}
    \end{subfigure}
    \begin{subfigure}[b]{0.24\linewidth}
        \centering
        \includegraphics[width=\linewidth]{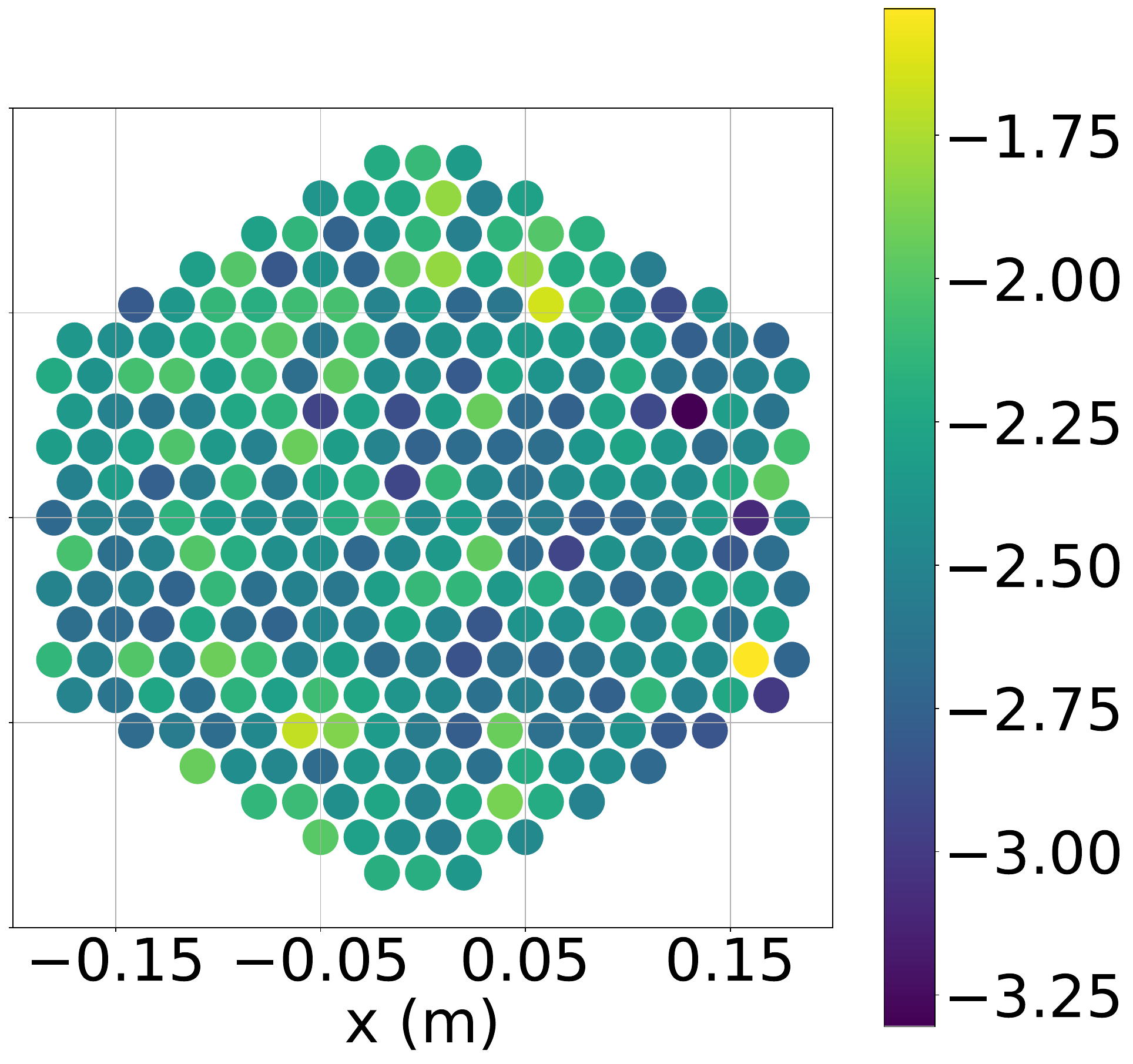}
        \caption{Standard Deviation}
        \label{fig:std_dev_abs_err_cropped_dataset}
    \end{subfigure}
    \caption{The distribution of data in the cropped 4-minute initial dataset, as well as the mean and standard deviation of the absolute errors (scaled by a base-10 logarithm) computed over the last 5 seconds of each RKL-SAC trial using the cropped dataset. The mean error across the 293 test points is 0.03 m, with an average standard deviation of 0.0237 m.}
    \label{fig:rkl_balance_ball_distribution_cropped_dataset}
\end{figure}

We also evaluate RKL-SAC on never-before-seen states by removing all data with positions in the first quadrant from a 4-minute dataset (Appx. Fig.~\ref{fig:rkl_balance_ball_distribution_cropped_dataset}). Using the polynomial observation function, RKL-SAC achieves a mean error of 0.03 m and a standard deviation of 0.0237 m across 293 test points. Online model updates quickly estimate the Koopman model in regions that is not presented in the dataset. In contrast, KL-SAC performs poorly with the same cropped dataset. Although RKL-SAC does not outperform than using the full dataset, it still exceeds RL-SAC, demonstrating strong stability and sample efficiency.

Again, online model updates are critical to RKL's strong performance. A KL-SAC controller trained on a larger initial dataset does not outperform an RKL-SAC controller trained on a smaller initial dataset with 20-second model updates, which means 20 seconds of online data collected during the process of achieving the control goal is more effective than several minutes of manually collected data, as shown in Fig.~\ref{fig:mean_abs_err}. This empirically validates the weak ACG hypothesis. Real-time model updates can efficiently utilize this data and enhance the controller's performance.

\subsubsection{Tracking of Trajectories Containing Contacts}

\begin{figure}[htbp]
    \centering
    \includegraphics[width=0.5\linewidth]{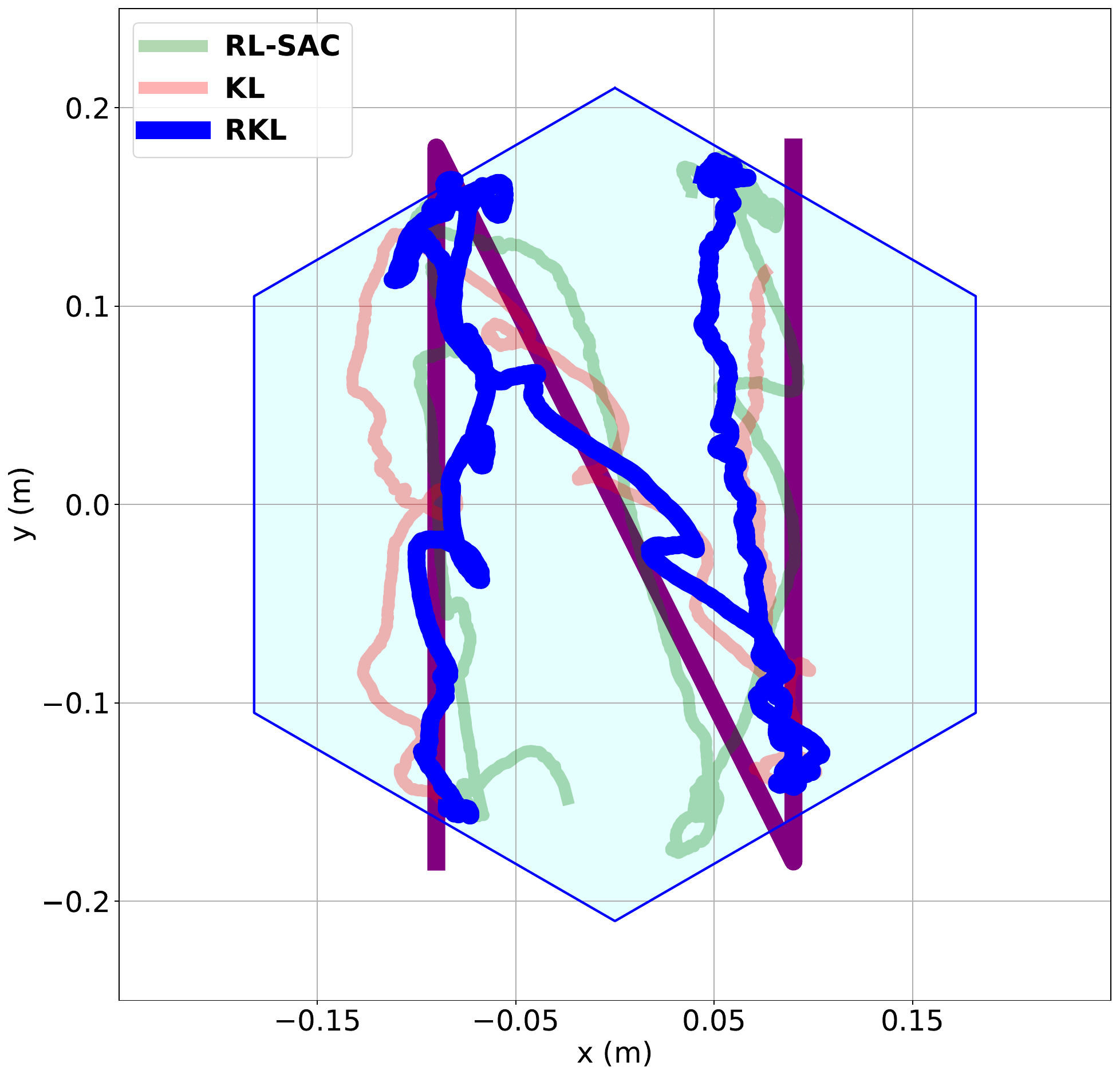}
    \caption{Tracking a reference trajectory containing contacts. The start, end and both corners of the reference trajectory are outside the platform boundary. The results shown are the best trials for each method (videos available). The cyan region denotes the platform area, which is slightly smaller than its actual size due to Apriltag measurement accuracy.}
    \label{fig:track_N}
\end{figure}

We extend our study to control the puck such that it follows a reference trajectory---one is shaped like the letter ``N'' and involves contacts with the boundaries (Appx.~Fig.~\ref{fig:track_N}). To encourage contacts, the start, end, and both corners of the ``N'' lie outside the platform's boundaries. The reference velocity for each segment of the ``N'' is constant, with each segment taking 7 seconds to complete.

\setlength{\intextsep}{0pt}
\begin{wrapfigure}{l}{0.59\textwidth}
    \centering
    \begin{tabular}{|c|c|c|c|c|}
        \hline
        Method & Mean (m) & Max (m) & Min (m) \\
        \hline
        KL-SAC (4 min) & 0.1079 & 0.2319 & 0.0628 \\
        RKL-SAC (4 min) & \textbf{0.0636} & \textbf{0.0812} & \textbf{0.0477} \\
        RL-SAC & 0.0821 & 0.1006 & 0.0732 \\
        \hline
    \end{tabular}
    \captionsetup{type=table}
    \caption{KL-SAC vs. RKL-SAC vs. RL-SAC in tracking an ``N'' trajectory.}
    \label{tab:track_N_performance}
\end{wrapfigure}

We conduct ten trials per method, starting from the lower-left corner of the ``N''. For each method, we select the five best trials based on the Fréchet distance between the reference and puck's position trajectories, and report the mean, maximum, and minimum value (Table~\ref{tab:track_N_performance}). The RKL-SAC and KL-SAC controllers use the a 4-minute dataset, while the RL-SAC policy is chosen based on the lowest mean error. RKL-SAC achieves the best performance. Trajectories from the best trials are shown in Appx. Fig.~\ref{fig:track_N}.

\section{Conclusion}\label{sec:conclusion}

In this paper, we present a sample-efficient Koopman-based learning pipeline, RKL. This pipeline can rapidly generate a controller capable of not only controlling hybrid nonlinear dynamical systems but also quickly learning and recursively updating models in real time. Our experiments on a simulated planar two-link arm and a soft-actuated hardware platform demonstrate the superior sample efficiency and improved control performance of RKL compared to SotA benchmarks. Additionally, we present the first formal convergence analysis of EDMD and RLS under continuous data growth within Markov chains, providing detailed sufficient conditions for convergence---critical for practical use and absent in prior work. We not only empirically validate the ACG hypothesis but also use our analysis results to explain it. Furthermore, we provide detailed algorithmic analyses of EDMD and RLS to prove that RKL is lightweight and fast, with complexity independent of dataset size.

\section{Limitations}\label{sec:limitations}

While model updating can significantly improve the control performance, we have observed cases where the RKL controller struggles to move the puck away from its current location during experiments on the SSP. This typically occurs when the puck remains in a small region for an extended period. As a result, the dataset becomes imbalanced: the model achieves high accuracy in predicting state transitions near the current location and at zero velocity but performs worse elsewhere. Additionally, because it is difficult to collect demonstration data on the SSP---whether from human operators or teaching policies---the initial model often ends up with an invariant measure that is far from the true distribution, as discussed in Sec.~\ref{sec:planar_two_link_arm_experiments}. Therefore, such an initial model is a poor starting point for RKL and can lead to the mentioned problems above.

Based on our convergence analysis, this issue can potentially be addressed by using high-quality demonstration data or by applying ergodicity-related metrics to guide the controller’s behavior while attempting the control goal. Many systems are naturally poorly suited for learning from demonstration, making ergodicity-related metrics especially important in those cases.

\section*{Acknowledgments}\label{sec:acknowledgments}

This work is supported in part by US Army Research Office (ARO) grant  W911NF-19-1-0233 and by a National Defense Science and Engineering Graduate Fellowship. In addition, we thank Jake Ketchum and Helena Young for their support with the hardware.

\clearpage

\bibliography{references}  

\begin{thebibliography}{49}
\providecommand{\natexlab}[1]{#1}
\providecommand{\url}[1]{\texttt{#1}}
\expandafter\ifx\csname urlstyle\endcsname\relax
  \providecommand{\doi}[1]{doi: #1}\else
  \providecommand{\doi}{doi: \begingroup \urlstyle{rm}\Url}\fi

\bibitem[Kleff et~al.(2021)Kleff, Meduri, Budhiraja, Mansard, and Righetti]{kleff2021}
S.~Kleff, A.~Meduri, R.~Budhiraja, N.~Mansard, and L.~Righetti.
\newblock High-frequency nonlinear model predictive control of a manipulator.
\newblock In \emph{2021 IEEE International Conference on Robotics and Automation (ICRA)}, pages 7330--7336, 2021.
\newblock \doi{10.1109/ICRA48506.2021.9560990}.

\bibitem[Grandia et~al.(2023)Grandia, Jenelten, Yang, Farshidian, and Hutter]{grandia2023}
R.~Grandia, F.~Jenelten, S.~Yang, F.~Farshidian, and M.~Hutter.
\newblock Perceptive locomotion through nonlinear model-predictive control.
\newblock \emph{IEEE Transactions on Robotics}, 39\penalty0 (5):\penalty0 3402--3421, 2023.
\newblock \doi{10.1109/TRO.2023.3275384}.

\bibitem[Meduri et~al.(2023)Meduri, Shah, Viereck, Khadiv, Havoutis, and Righetti]{meduri2023}
A.~Meduri, P.~Shah, J.~Viereck, M.~Khadiv, I.~Havoutis, and L.~Righetti.
\newblock {BiConMP}: A nonlinear model predictive control framework for whole body motion planning.
\newblock \emph{IEEE Transactions on Robotics}, 39\penalty0 (2):\penalty0 905--922, 2023.
\newblock \doi{10.1109/TRO.2022.3228390}.

\bibitem[Le~Cleac'h et~al.(2024)Le~Cleac'h, Howell, Yang, Lee, Zhang, Bishop, Schwager, and Manchester]{le2024}
S.~Le~Cleac'h, T.~A. Howell, S.~Yang, C.-Y. Lee, J.~Zhang, A.~Bishop, M.~Schwager, and Z.~Manchester.
\newblock Fast contact-implicit model predictive control.
\newblock \emph{IEEE Transactions on Robotics}, 40:\penalty0 1617--1629, 2024.
\newblock \doi{10.1109/TRO.2024.3351554}.

\bibitem[Schulman et~al.(2017)Schulman, Wolski, Dhariwal, Radford, and Klimov]{schulman2017}
J.~Schulman, F.~Wolski, P.~Dhariwal, A.~Radford, and O.~Klimov.
\newblock Proximal policy optimization algorithms.
\newblock \emph{CoRR}, abs/1707.06347, 2017.
\newblock URL \url{http://arxiv.org/abs/1707.06347}.

\bibitem[Fujimoto et~al.(2018)Fujimoto, van Hoof, and Meger]{fujimoto2018}
S.~Fujimoto, H.~van Hoof, and D.~Meger.
\newblock Addressing function approximation error in actor-critic methods.
\newblock In J.~Dy and A.~Krause, editors, \emph{Proceedings of the 35th International Conference on Machine Learning}, volume~80 of \emph{Proceedings of Machine Learning Research}, pages 1587--1596. PMLR, 10--15 Jul 2018.
\newblock URL \url{https://proceedings.mlr.press/v80/fujimoto18a.html}.

\bibitem[Haarnoja et~al.(2018)Haarnoja, Zhou, Abbeel, and Levine]{haarnoja2018}
T.~Haarnoja, A.~Zhou, P.~Abbeel, and S.~Levine.
\newblock Soft actor-critic: Off-policy maximum entropy deep reinforcement learning with a stochastic actor.
\newblock In \emph{Proceedings of the 35th International Conference on Machine Learning}, volume~80 of \emph{Proceedings of Machine Learning Research}, pages 1861--1870. PMLR, 10--15 Jul 2018.
\newblock URL \url{https://proceedings.mlr.press/v80/haarnoja18b.html}.

\bibitem[Jitosho et~al.(2023)Jitosho, Lum, Okamura, and Liu]{jitosho2023}
R.~Jitosho, T.~G.~W. Lum, A.~Okamura, and K.~Liu.
\newblock Reinforcement learning enables real-time planning and control of agile maneuvers for soft robot arms.
\newblock In \emph{Proceedings of The 7th Conference on Robot Learning}, volume 229 of \emph{Proceedings of Machine Learning Research}, pages 1131--1153. PMLR, 06--09 Nov 2023.

\bibitem[Rajeswaran et~al.(2018)Rajeswaran, Kumar, Gupta, Vezzani, Schulman, Todorov, and Levine]{rajeswaran2018}
A.~Rajeswaran, V.~Kumar, A.~Gupta, G.~Vezzani, J.~Schulman, E.~Todorov, and S.~Levine.
\newblock Learning complex dexterous manipulation with deep reinforcement learning and demonstrations.
\newblock In \emph{Proceedings of Robotics: Science and Systems}, Pittsburgh, Pennsylvania, June 2018.
\newblock \doi{10.15607/RSS.2018.XIV.049}.

\bibitem[Chen et~al.(2022)Chen, Wu, Wang, Feng, Jiang, Lu, McAleer, Dong, Zhu, and Yang]{chen2022}
Y.~Chen, T.~Wu, S.~Wang, X.~Feng, J.~Jiang, Z.~Lu, S.~McAleer, H.~Dong, S.-C. Zhu, and Y.~Yang.
\newblock Towards human-level bimanual dexterous manipulation with reinforcement learning.
\newblock In S.~Koyejo, S.~Mohamed, A.~Agarwal, D.~Belgrave, K.~Cho, and A.~Oh, editors, \emph{Advances in Neural Information Processing Systems}, volume~35, pages 5150--5163. Curran Associates, Inc., 2022.
\newblock URL \url{https://proceedings.neurips.cc/paper_files/paper/2022/file/217a2a387f52c30755c37b0a73430291-Paper-Datasets_and_Benchmarks.pdf}.

\bibitem[Williams et~al.(2017)Williams, Wagener, Goldfain, Drews, Rehg, Boots, and Theodorou]{williams2017}
G.~Williams, N.~Wagener, B.~Goldfain, P.~Drews, J.~M. Rehg, B.~Boots, and E.~A. Theodorou.
\newblock Information theoretic mpc for model-based reinforcement learning.
\newblock In \emph{2017 IEEE International Conference on Robotics and Automation (ICRA)}, pages 1714--1721, 2017.
\newblock \doi{10.1109/ICRA.2017.7989202}.

\bibitem[Lee et~al.(2020)Lee, Hwangbo, Wellhausen, Koltun, and Hutter]{joonho2020}
J.~Lee, J.~Hwangbo, L.~Wellhausen, V.~Koltun, and M.~Hutter.
\newblock Learning quadrupedal locomotion over challenging terrain.
\newblock \emph{Science Robotics}, 5\penalty0 (47):\penalty0 eabc5986, 2020.
\newblock \doi{10.1126/scirobotics.abc5986}.

\bibitem[Mock and Muknahallipatna(2023)]{mock2023}
J.~W. Mock and S.~S. Muknahallipatna.
\newblock A comparison of {PPO}, {TD3} and {SAC} reinforcement algorithms for quadruped walking gait generation.
\newblock \emph{Journal of Intelligent Learning Systems and Applications}, 15\penalty0 (1):\penalty0 36--56, 2023.
\newblock \doi{10.4236/jilsa.2023.151003}.

\bibitem[Tu(2013)]{tu2013}
J.~H. Tu.
\newblock \emph{Dynamic mode decomposition: Theory and applications}.
\newblock PhD thesis, Princeton University, 2013.
\newblock URL \url{http://turing.library.northwestern.edu/login?url=https://www.proquest.com/dissertations-theses/dynamic-mode-decomposition-theory-applications/docview/1458341928/se-2}.

\bibitem[Williams et~al.(2015)Williams, Kevrekidis, and Rowley]{williams2015}
M.~O. Williams, I.~G. Kevrekidis, and C.~W. Rowley.
\newblock A data--driven approximation of the {Koopman} operator: Extending dynamic mode decomposition.
\newblock \emph{Journal of Nonlinear Science}, 25:\penalty0 1307--1346, 2015.
\newblock \doi{10.1007/s00332-015-9258-5}.

\bibitem[Rowley and Dawson(2017)]{rowley2017}
C.~W. Rowley and S.~T. Dawson.
\newblock Model reduction for flow analysis and control.
\newblock \emph{Annual Review of Fluid Mechanics}, 49\penalty0 (49):\penalty0 387--417, 2017.
\newblock ISSN 1545-4479.
\newblock \doi{10.1146/annurev-fluid-010816-060042}.

\bibitem[Geneva and Zabaras(2022)]{geneva2022}
N.~Geneva and N.~Zabaras.
\newblock Transformers for modeling physical systems.
\newblock \emph{Neural Networks}, 146:\penalty0 272--289, 2022.
\newblock ISSN 0893-6080.
\newblock \doi{10.1016/j.neunet.2021.11.022}.

\bibitem[Susuki and Mezić(2010)]{susuki2010}
Y.~Susuki and I.~Mezić.
\newblock Nonlinear {Koopman} modes of coupled swing dynamics and coherency identification.
\newblock In \emph{IEEE PES General Meeting}, pages 1--8, 2010.
\newblock \doi{10.1109/PES.2010.5589363}.

\bibitem[Bruder et~al.(2019)Bruder, Gillespie, Remy, and Vasudevan]{bruder2019}
D.~Bruder, B.~Gillespie, C.~D. Remy, and R.~Vasudevan.
\newblock Modeling and control of soft robots using the {Koopman} operator and model predictive control.
\newblock In \emph{Proceedings of Robotics: Science and Systems}, Freiburg im Breisgau, Germany, June 2019.
\newblock \doi{10.15607/RSS.2019.XV.060}.

\bibitem[Bruder et~al.(2024)Bruder, Bombara, and Wood]{bruder2024}
D.~Bruder, D.~Bombara, and R.~J. Wood.
\newblock A {Koopman}-based residual modeling approach for the control of a soft robot arm.
\newblock \emph{The International Journal of Robotics Research}, 0\penalty0 (0), 2024.
\newblock \doi{10.1177/02783649241272114}.

\bibitem[Abraham and Murphey(2019)]{Abaraham2019}
I.~Abraham and T.~D. Murphey.
\newblock Active learning of dynamics for data-driven control using {Koopman} operators.
\newblock \emph{IEEE Transactions on Robotics}, 35\penalty0 (5):\penalty0 1071--1083, 2019.
\newblock \doi{10.1109/TRO.2019.2923880}.

\bibitem[Mamakoukas et~al.(2021)Mamakoukas, Castaño, Tan, and Murphey]{mamakoukas2021}
G.~Mamakoukas, M.~L. Castaño, X.~Tan, and T.~D. Murphey.
\newblock Derivative-based {Koopman} operators for real-time control of robotic systems.
\newblock \emph{IEEE Transactions on Robotics}, 37\penalty0 (6):\penalty0 2173--2192, 2021.
\newblock \doi{10.1109/TRO.2021.3076581}.

\bibitem[Li et~al.(2025)Li, Abuduweili, Sun, Chen, Zhao, and Liu]{li2024}
F.~Li, A.~Abuduweili, Y.~Sun, R.~Chen, W.~Zhao, and C.~Liu.
\newblock Continual learning and lifting of {Koopman} dynamics for linear control of legged robots.
\newblock In \emph{Proceedings of the 7th Annual Learning for Dynamics and Control Conference}, volume 283 of \emph{Proceedings of Machine Learning Research}, pages 1--30. PMLR, 4--6 June 2025.

\bibitem[Pomerleau(1988)]{pomerleau1988}
D.~A. Pomerleau.
\newblock Alvinn: An autonomous land vehicle in a neural network.
\newblock In D.~Touretzky, editor, \emph{Advances in Neural Information Processing Systems}, volume~1. Morgan-Kaufmann, 1988.
\newblock URL \url{https://proceedings.neurips.cc/paper_files/paper/1988/file/812b4ba287f5ee0bc9d43bbf5bbe87fb-Paper.pdf}.

\bibitem[Hayes(1996)]{hayes1996}
M.~H. Hayes.
\newblock Chapter 9.4 - recursive least squares.
\newblock In \emph{Statistical Digital Signal Processing and Modeling}. John Wiley \& Sons, 1996.
\newblock \doi{10.1080/00401706.1997.10485128}.

\bibitem[Nagabandi et~al.(2020)Nagabandi, Konolige, Levine, and Kumar]{nagabandi2020}
A.~Nagabandi, K.~Konolige, S.~Levine, and V.~Kumar.
\newblock Deep dynamics models for learning dexterous manipulation.
\newblock In \emph{2020 Conference on Robot Learning (CoRL)}, pages 1101--1112. PMLR, 30 Oct--01 Nov 2020.
\newblock URL \url{https://proceedings.mlr.press/v100/nagabandi20a.html}.

\bibitem[Zhang et~al.(2019)Zhang, Rowley, Deem, and Cattafesta]{zhang2019}
H.~Zhang, C.~W. Rowley, E.~A. Deem, and L.~N. Cattafesta.
\newblock Online dynamic mode decomposition for time-varying systems.
\newblock \emph{SIAM Journal on Applied Dynamical Systems}, 18\penalty0 (3):\penalty0 1586--1609, 2019.
\newblock \doi{10.1137/18M1192329}.

\bibitem[Calderón et~al.(2021)Calderón, Schulz, Oehlschlägel, and Werner]{calderon2021}
H.~M. Calderón, E.~Schulz, T.~Oehlschlägel, and H.~Werner.
\newblock {Koopman} operator-based model predictive control with recursive online update.
\newblock In \emph{2021 European Control Conference (ECC)}, pages 1543--1549, 2021.
\newblock \doi{10.23919/ECC54610.2021.9655220}.

\bibitem[N{\"u}ske et~al.(2023)N{\"u}ske, Peitz, Philipp, Schaller, and Worthmann]{nuske2023}
F.~N{\"u}ske, S.~Peitz, F.~Philipp, M.~Schaller, and K.~Worthmann.
\newblock Finite-data error bounds for {Koopman}-based prediction and control.
\newblock \emph{Journal of Nonlinear Science}, 33\penalty0 (1):\penalty0 14, 2023.
\newblock \doi{10.1007/s00332-022-09862-1}.

\bibitem[Zhang and Zuazua(2023)]{zhang2023}
C.~Zhang and E.~Zuazua.
\newblock A quantitative analysis of {Koopman} operator methods for system identification and predictions.
\newblock \emph{Comptes Rendus. M{\'e}canique}, 351\penalty0 (S1):\penalty0 1--31, 2023.
\newblock \doi{10.5802/crmeca.138}.

\bibitem[Philipp et~al.(2024)Philipp, Schaller, Boshoff, Peitz, N{\"u}ske, and Worthmann]{philipp2024}
F.~M. Philipp, M.~Schaller, S.~Boshoff, S.~Peitz, F.~N{\"u}ske, and K.~Worthmann.
\newblock Extended dynamic mode decomposition: Sharp bounds on the sample efficiency.
\newblock \emph{arXiv preprint arXiv:2402.02494}, 2024.
\newblock URL \url{https://arxiv.org/abs/2402.02494}.

\bibitem[Breiman(1960)]{breiman1960}
L.~Breiman.
\newblock The strong law of large numbers for a class of {Markov} chains.
\newblock \emph{The Annals of Mathematical Statistics}, 31\penalty0 (3):\penalty0 801--803, 1960.
\newblock URL \url{https://www.jstor.org/stable/2237593}.

\bibitem[Koopman(1931)]{koopman1931}
B.~O. Koopman.
\newblock Hamiltonian systems and transformation in hilbert space.
\newblock \emph{Proceedings of the National Academy of Sciences}, 17\penalty0 (5):\penalty0 315--318, 1931.
\newblock \doi{10.1073/pnas.17.5.315}.

\bibitem[Asada(2023)]{asada2023}
H.~H. Asada.
\newblock Global, unified representation of heterogenous robot dynamics using composition operators: A {Koopman} direct encoding method.
\newblock \emph{IEEE/ASME Transactions on Mechatronics}, 28\penalty0 (5):\penalty0 2633--2644, 2023.
\newblock \doi{10.1109/TMECH.2023.3253599}.

\bibitem[Proctor et~al.(2016)Proctor, Brunton, and Kutz]{proctor2016}
J.~L. Proctor, S.~L. Brunton, and J.~N. Kutz.
\newblock Dynamic mode decomposition with control.
\newblock \emph{SIAM Journal on Applied Dynamical Systems}, 15\penalty0 (1):\penalty0 142--161, 2016.
\newblock \doi{10.1137/15M1013857}.

\bibitem[Sherman and Morrison(1950)]{sherman1950}
J.~Sherman and W.~J. Morrison.
\newblock Adjustment of an inverse matrix corresponding to a change in one element of a given matrix.
\newblock \emph{The Annals of Mathematical Statistics}, 21\penalty0 (1):\penalty0 124--127, 1950.
\newblock ISSN 00034851.
\newblock URL \url{http://www.jstor.org/stable/2236561}.

\bibitem[Ansari and Murphey(2016)]{Ansari2016}
A.~R. Ansari and T.~D. Murphey.
\newblock Sequential action control: Closed-form optimal control for nonlinear and nonsmooth systems.
\newblock \emph{IEEE Transactions on Robotics}, 32\penalty0 (5):\penalty0 1196--1214, 2016.
\newblock \doi{10.1109/TRO.2016.2596768}.

\bibitem[Nishimura and Schwager(2021)]{nishimura2021}
H.~Nishimura and M.~Schwager.
\newblock {SACBP}: Belief space planning for continuous-time dynamical systems via stochastic sequential action control.
\newblock \emph{The International Journal of Robotics Research}, 40\penalty0 (10-11):\penalty0 1167--1195, 2021.
\newblock \doi{10.1177/02783649211037697}.

\bibitem[Ketchum et~al.(2025)Ketchum, Avtges, Schlafly, Young, Kim, Truby, and Murphey]{ketchum2025}
J.~Ketchum, J.~Avtges, M.~Schlafly, H.~Young, T.~Kim, R.~L. Truby, and T.~D. Murphey.
\newblock Force and speed in a soft stewart platform, 2025.
\newblock URL \url{https://arxiv.org/abs/2504.13127}.

\bibitem[Kim et~al.(2024)Kim, Kaarthik, and Truby]{kim2024}
T.~Kim, P.~Kaarthik, and R.~L. Truby.
\newblock A flexible, architected soft robotic actuator for motorized extensional motion.
\newblock \emph{Advanced Intelligent Systems}, 6\penalty0 (11):\penalty0 2300866, 2024.
\newblock \doi{10.1002/aisy.202300866}.

\bibitem[Abraham et~al.(2017)Abraham, de~la Torre, and Murphey]{Abraham2017}
I.~Abraham, G.~de~la Torre, and T.~Murphey.
\newblock Model-based control using {Koopman} operators.
\newblock In \emph{Proceedings of Robotics: Science and Systems}, Cambridge, Massachusetts, July 2017.
\newblock \doi{10.15607/RSS.2017.XIII.052}.

\bibitem[Chen et~al.(2021)Chen, Wang, Zhou, and Ross]{chen2021}
X.~Chen, C.~Wang, Z.~Zhou, and K.~W. Ross.
\newblock Randomized ensembled double q-learning: Learning fast without a model.
\newblock In \emph{International Conference on Learning Representations}, 2021.
\newblock URL \url{https://openreview.net/forum?id=AY8zfZm0tDd}.

\bibitem[Avtges et~al.(2025)Avtges, Ketchum, Schlafly, Young, Kim, Pinosky, Truby, and Murphey]{avtges2025}
J.~Avtges, J.~Ketchum, M.~Schlafly, H.~Young, T.~Kim, A.~Pinosky, R.~L. Truby, and T.~D. Murphey.
\newblock Real-time reinforcement learning for dynamic tasks with a parallel soft robot.
\newblock In \emph{2025 IEEE/RSJ International Conference on Intelligent Robots and Systems (IROS)}, 2025.

\bibitem[Boyd and Vandenberghe(2018)]{boyd2018}
S.~Boyd and L.~Vandenberghe.
\newblock Part {III} least squares.
\newblock In \emph{Introduction to applied linear algebra: vectors, matrices, and least squares}. Cambridge university press, 2018.
\newblock URL \url{https://web.stanford.edu/~boyd/vmls/vmls.pdf}.

\bibitem[Stewart(1969)]{stewart1969}
G.~W. Stewart.
\newblock On the continuity of the generalized inverse.
\newblock \emph{SIAM Journal on Applied Mathematics}, 17\penalty0 (1):\penalty0 33--45, 1969.
\newblock ISSN 00361399.
\newblock URL \url{http://www.jstor.org/stable/2099241}.

\bibitem[Foundation()]{gymnasium_reacher}
F.~Foundation.
\newblock Reacher - gymnasium documentation.
\newblock \url{https://gymnasium.farama.org/environments/mujoco/reacher/}.

\bibitem[Berrueta et~al.(2024)Berrueta, Pinosky, and Murphey]{berrueta2024}
T.~A. Berrueta, A.~Pinosky, and T.~D. Murphey.
\newblock Maximum diffusion reinforcement learning.
\newblock \emph{Nature Machine Intelligence}, 6\penalty0 (5):\penalty0 504--514, 2024.

\bibitem[Raffin et~al.(2021)Raffin, Hill, Gleave, Kanervisto, Ernestus, and Dormann]{stable-baselines3}
A.~Raffin, A.~Hill, A.~Gleave, A.~Kanervisto, M.~Ernestus, and N.~Dormann.
\newblock Stable-baselines3: Reliable reinforcement learning implementations.
\newblock \emph{Journal of Machine Learning Research}, 22\penalty0 (268):\penalty0 1--8, 2021.
\newblock URL \url{http://jmlr.org/papers/v22/20-1364.html}.

\bibitem[Williams et~al.(2016)Williams, Drews, Goldfain, Rehg, and Theodorou]{williams_aggressive_2016}
G.~Williams, P.~Drews, B.~Goldfain, J.~M. Rehg, and E.~A. Theodorou.
\newblock Aggressive driving with model predictive path integral control.
\newblock In \emph{2016 {IEEE} {International} {Conference} on {Robotics} and {Automation} ({ICRA})}, pages 1433--1440, 2016.
\newblock \doi{10.1109/ICRA.2016.7487277}.

\end{thebibliography}

\appendices
\section{Sequential Action Control}\label{sec:mpc-sac}

A continuous-time MPC policy determines control inputs for a dynamical system by repeatedly solving a constrained trajectory optimization problem. At each iteration, we solve
\begin{equation}\label{eq:mpc_traj_opt}
    \begin{split}
        \min_{\mathbf{x}\left(\cdot\right),\mathbf{u}\left(\cdot\right)}\hspace{0.3cm} & J= \underbrace{m\left(\mathbf{x}\left(T\right)\right)}_{\text{terminal cost}} + \int_{0}^{T}{\underbrace{l\left(\mathbf{x}\left(\tau\right),\mathbf{u}\left(\tau\right)\right)}_{\text{running cost}}d\tau}\\        
        \text{s.t.}\hspace{0.3cm} & \mathbf{x}\left(0\right)=\mathbf{x}_{\text{init}},\text{ initial condition},\\
        & \dot{\mathbf{x}}\left(t\right) = \mathbf{f}(\mathbf{x}\left(t\right),\mathbf{u}\left(t\right)),\text{ dynamics},\\
        & \mathbf{c}_{\text{eq}}(\mathbf{x}\left(t\right),\mathbf{u}\left(t\right))\leq\mathbf{0},\text{ equality constraints},\\
        & \mathbf{c}_{\text{iq}}(\mathbf{x}\left(t\right),\mathbf{u}\left(t\right))=\mathbf{0},\text{ inequality constraints}.
    \end{split}
\end{equation}
Once the optimal control sequence is obtained, only the first control input is applied to the system, and the process is repeated at the next time step.

Sequential Action Control (MPC-SAC)~\citep{Ansari2016} is a model-based approach that solves MPC in real time within a closed-loop framework for nonlinear systems. Instead of iteratively optimizing finite-horizon control sequences to minimize a given objective, MPC-SAC utilizes a closed-form expression to determine individual control actions. These actions, applied over short durations, are designed to optimally enhance a tracking objective over a long time horizon. Moreover, the existence and uniqueness of optimal actions are globally guaranteed. Here, we introduce the algorithm in the context of Koopman-based control, and the detailed derivation is presented in~\citep{Ansari2016,Abaraham2019}. The algorithm is summarized in Alg.~\ref{alg:mpc-sac}.

\setlength{\intextsep}{\intextsepdefault}
\begin{algorithm}[H]\caption{MPC-SAC using a Koopman model}\label{alg:mpc-sac}
    \begin{algorithmic}
        \Require Nominal policy $\boldsymbol{\mu}$, weight matrix $\bar{\mathbf{R}}$, Koopman matrix $\mathbf{K}$, current feedback $\mathbf{z}$, reference $\mathbf{z}_{\text{ref}}$ and $\mathbf{u}_{\text{ref}}$, time step length $\delta t$, horizon length $H$
        \State $\mathbf{u}^*=\text{MPC-SAC}\left(\mathbf{K}, \mathbf{z}, \mathbf{z}_{\text{ref}}, \mathbf{u}_{\text{ref}}, \boldsymbol{\mu}, \bar{\mathbf{R}}, \delta t, H\right)$
        \State $\quad\boldsymbol{\mu}\text{.update}\left(\mathbf{K}, \mathbf{z}_{\text{ref}}, \mathbf{u}_{\text{ref}}\right)$
        \State $\quad\text{Adjoint.reset}\left(\delta t, H\right)$
        \State $\quad \text{Objective.reset}\left(\mathbf{z}_\text{ref}, \mathbf{u}_\text{ref}\right)$
	\State $\quad\text{traj}=\boldsymbol{\mu}\text{.simulate}\left(H, \delta t,\text{Objective}\right)$
	\State $\quad\boldsymbol{\rho}\text{\_traj=Adjoint.simulate}\left(\text{traj}\right)$
        \State $\quad\mathbf{u}=\boldsymbol{\mu}\text{.solve}\left(\mathbf{z},\text{Objective}\right)$
	\State $\quad\mathbf{u}^*=-\bar{\mathbf{R}}^{-1}\mathbf{K}_u^\top\boldsymbol{\rho}\left(0\right)+\mathbf{u}$
    \end{algorithmic}
\end{algorithm}

The dynamics model used to synthesize the controller is a continuous-time Koopman model,
\begin{equation}
    \dot{\mathbf{z}}\left(t\right)=\mathbf{f}\left(\mathbf{z}\left(t\right),\mathbf{u}\left(t\right)\right)=\mathbf{K}_z\mathbf{z}\left(t\right)+\mathbf{K}_u\mathbf{u}\left(t\right),
\end{equation}
where $\mathbf{z}$ is the observation of the state.
Define the objective function $J_1\in\mathbb{R}$,
\begin{equation}\label{eq:sac objective}
    J_1=\int_{t_0}^{t_0+T}{l\left(\mathbf{z}\left(s\right),\boldsymbol{\mu}\left(\mathbf{z}\left(s\right)\right)\right)ds}+m\left(\mathbf{z}\left(t_0+T\right)\right),
\end{equation}
where $\boldsymbol{\mu}\left(\mathbf{z}\right)$ is a nominal feedback control policy. We use LQR as the nominal policy in our experiments.

MPC-SAC finds the change in the objective (Eq.~\ref{eq:sac objective}) due to inserting a new control policy $\boldsymbol{\mu}^*$ into the nominal trajectory for a short duration around $t=\tau\in\left[t_0, t_0+T\right]$. The related mode insertion gradient is given by
\begin{equation}\label{eq:mode insertion gradient}
    \frac{\partial J_1}{\partial\lambda}|_{t=\tau,\lambda=0} = \boldsymbol{\rho}\left(\tau\right)^\top\left(\mathbf{f}_2-\mathbf{f}_1\right),
\end{equation}
where $\mathbf{f}_1=\mathbf{f}\left(\mathbf{z}\left(\tau\right),\boldsymbol{\mu}\left(\mathbf{z}\left(\tau\right)\right)\right)$ and $\mathbf{f}_2=\mathbf{f}\left(\mathbf{z}\left(\tau\right),\boldsymbol{\mu}^*\left(\mathbf{z}\left(\tau\right)\right)\right)$. The adjoint variable (or costate) $\boldsymbol{\rho}$ satisfies
\begin{equation}\label{eq:sac rho}
    \dot{\boldsymbol{\rho}} = -\left(\frac{\partial l}{\partial \mathbf{z}} + \frac{\partial\boldsymbol{\mu}}{\partial \mathbf{z}}^\top\frac{\partial l}{\partial \mathbf{u}}\right) - \left(\frac{\partial \mathbf{f}}{\partial \mathbf{z}} + \frac{\partial \mathbf{f}}{\partial \mathbf{u}}\frac{\partial\boldsymbol{\mu}}{\partial \mathbf{z}}\right)^\top\boldsymbol{\rho},
\end{equation}
with the terminal condition
\begin{equation}\label{eq:rho terminal condition}
    \boldsymbol{\rho}\left(t_0+T\right) = \frac{\partial}{\partial \mathbf{z}}m\left(\mathbf{z}\left(t_0 + T\right)\right).
\end{equation}
We first use the nominal policy to simulate the system's trajectory over the horizon length, starting from the current state. Then, we compute the trajectory of $\boldsymbol{\rho}$ backwards from the terminal condition.

Finally, MPC-SAC solves a new optimization problem to minimize the mode insertion gradient,
\begin{equation}
    J_2=\int_{t_0}^{t_0+T}\frac{\partial J_1}{\partial\lambda}|_{t=s,\lambda=0}+\frac{1}{2}\left\|\boldsymbol{\mu}^*\left(s\right)-\boldsymbol{\mu}\left(\mathbf{z}\left(t\right)\right)\right\|_{\bar{R}}^2ds,
\end{equation}
where $\bar{\mathbf{R}}$ is a diagonal matrix to bound the change from $\boldsymbol{\mu}$ to $\boldsymbol{\mu}^*$. It can be shown that the analytical solution is~\citep{Ansari2016,Abaraham2019}
\begin{equation}
    \boldsymbol{\mu}^*\left(t\right)=-\bar{\mathbf{R}}^{-1}\mathbf{K}_u^\top\boldsymbol{\rho}\left(t\right)+\boldsymbol{\mu}\left(\mathbf{z}\left(t\right)\right).
\end{equation}
Finally, $\boldsymbol{\mu}^*\left(0\right)$ is applied to the system.

\section{RLS Update is Equivalent to EDMD}\label{sec:rls=edmd}

\begin{algorithm}[H]\caption{Online Model Update using RLS}\label{alg:rls}
    \begin{algorithmic}
        \State $\mathbf{K}_{\text{new}}, \mathbf{P}_{\text{new}}$ = RLS($\mathbf{K}$, $\mathbf{P}$, $\mathbf{z}$, $\mathbf{u}$, $\bar{\mathbf{z}}$)
            \State \quad$\boldsymbol{\alpha}=\left[\mathbf{z};\mathbf{u}\right]$\Comment{column vector}
            \State \quad$\boldsymbol{\beta}=\left[\bar{\mathbf{z}};\mathbf{u}\right]$\Comment{column vector}
            \State \quad$\gamma=1/\left(1+\boldsymbol{\alpha}^\top\mathbf{P}\boldsymbol{\alpha}\right)$
            \State \quad$\mathbf{P}_{\text{new}}=\mathbf{P}-\gamma\mathbf{P}\boldsymbol{\alpha}\boldsymbol{\alpha}^\top\mathbf{P}$
            \State \quad$\mathbf{K}_{\text{new}}=\mathbf{K}+\gamma\left(\boldsymbol{\beta}-\mathbf{K}\boldsymbol{\alpha}\right)\boldsymbol{\alpha}^\top\mathbf{P}$
    \end{algorithmic}
\end{algorithm}

Here we prove that the RLS update algorithm for Koopman models (Alg.~\ref{alg:rls}) is mathematically equivalent to retraining a Koopman model using EDMD. Similar proofs are also presented by~\citep{zhang2019,calderon2021}.

Consider the two data matrices used in EDMD. At time step $k$, assume that we have data matrices $\mathbf{Y}_k$ and $\bar{\mathbf{Y}}_k$, and the latest estimate of the Koopman matrix $\mathbf{K}_k$. In addition, we define
\begin{equation}\label{eq:Q_P_def}
    \mathbf{Q}_k:=\bar{\mathbf{Y}}_k\mathbf{Y}_k^\top, \mathbf{P}_k:=\left(\mathbf{Y}_k\mathbf{Y}_k^\top\right)^\dagger.
\end{equation}
Assume that we want to update the model after one time step. We got a new pair of data $\left\{\boldsymbol{\alpha}_k,\boldsymbol{\beta}_k\right\}$, and the data matrices becomes
\begin{equation}
    \mathbf{Y}_{k+1}=
    \begin{bmatrix}
        \mathbf{Y}_k & \boldsymbol{\alpha}_{k}
    \end{bmatrix},
\end{equation}
\begin{equation}
    \bar{\mathbf{Y}}_{k+1}=
    \begin{bmatrix}
        \bar{\mathbf{Y}}_k & \boldsymbol{\beta}_{k}
    \end{bmatrix}.
\end{equation}
Now we can calculate the new $\mathbf{Q}$ and $\mathbf{P}^\dagger$
\begin{equation}\label{eq:Q_{k+1}}
    \mathbf{Q}_{k+1}=
    \begin{bmatrix}
        \bar{\mathbf{Y}}_k & \boldsymbol{\beta}_{k}
    \end{bmatrix}
    \begin{bmatrix}
        \mathbf{Y}_k^\top \\ \boldsymbol{\alpha}_{k}^\top
    \end{bmatrix}=
    \bar{\mathbf{Y}}_k\mathbf{Y}_k^\top+\boldsymbol{\beta}_{k}\boldsymbol{\alpha}_{k}^\top,
\end{equation}
\begin{equation}\label{eq:P_{k+1}}
    \mathbf{P}_{k+1}^\dagger=
    \begin{bmatrix}
        \mathbf{Y}_k & \boldsymbol{\alpha}_k
    \end{bmatrix}
    \begin{bmatrix}
        \mathbf{Y}_k^\top \\ \boldsymbol{\alpha}_k^\top
    \end{bmatrix}=
    \mathbf{Y}_k\mathbf{Y}_k^\top+\boldsymbol{\alpha}_k\boldsymbol{\alpha}_k^\top.
\end{equation}

Now we want to update the model. Apply the EDMD update, we get
\begin{equation}\label{eq:edmd 2}
    \mathbf{K}_{k+1}=\mathbf{Q}_{k+1}\mathbf{P}_{k+1}.
\end{equation}
However, computing this needs computing the inverse of a large matrix, and involves several large matrix multiplications. To avoid this, first recall the Sherman-Morrison formula~\citep{sherman1950}. Suppose $\mathbf{A}$ is an invertible square real matrix, and $\mathbf{u}$, $\mathbf{v}$ are column real vectors. Then $\mathbf{A}+\mathbf{u}\mathbf{v}^\top$ is invertible if and only if $1+\mathbf{v}^\top\mathbf{A}^{-1}\mathbf{u}\neq 0$. In this case,
\begin{equation}
    \left(\mathbf{A}+\mathbf{u}\mathbf{v}^\top\right)^{-1}=\mathbf{A}^{-1}-\frac{\mathbf{A}^{-1}\mathbf{u}\mathbf{v}^\top\mathbf{A}^{-1}}{1+\mathbf{v}^\top\mathbf{A}^{-1}\mathbf{u}}.
\end{equation}
This formula can be used to compute $\mathbf{P}_{k+1}$. Start from Eq.~\ref{eq:P_{k+1}},
\begin{equation}\label{eq:P_{k+1} update}
    \begin{split}
        \mathbf{P}_{k+1}&=\left(\mathbf{P}_k^\dagger+\boldsymbol{\alpha}_k\boldsymbol{\alpha}_k^\top\right)^{-1}\\
        &=\mathbf{P}_k-\frac{\mathbf{P}_k\boldsymbol{\alpha}_k\boldsymbol{\alpha}_k^\top\mathbf{P}_k}{1+\boldsymbol{\alpha}_k^\top\mathbf{P}_k\boldsymbol{\alpha}_k}\\
        &=\mathbf{P}_k-\gamma_k\mathbf{P}_k\boldsymbol{\alpha}_k\boldsymbol{\alpha}_k^\top\mathbf{P}_k,
    \end{split}
\end{equation}
\begin{equation}
    \gamma_k:=\frac{1}{1+\boldsymbol{\alpha}_k^\top\mathbf{P}_k\boldsymbol{\alpha}_k}.
\end{equation}
By substituting Eq.~\ref{eq:Q_{k+1}} and~\ref{eq:P_{k+1} update} into Eq.~\ref{eq:edmd 2} and simplifing the expression, we can update $\mathbf{K}_k$ without calculating any matrix inverses,
\begin{equation}\label{eq:K_{k+1} update}
    \mathbf{K}_{k+1}=\mathbf{K}_k+\gamma_k\left(\boldsymbol{\beta}_k-\mathbf{K}_k\boldsymbol{\alpha}_k\right)\boldsymbol{\alpha}_k^\top\mathbf{P}_k.
\end{equation}
Note that we require $\mathbf{P}$ to be always full rank.

\section{Algorithmic Complexity Analysis}\label{sec:algorithmic_complexity_analysis}

\subsection{EDMD Complexity}\label{sec:edmd_complexity}

After collecting $N+1$ data pairs $\left\{\mathbf{x}, \mathbf{u}\right\}$ , let $\mathbf{Y}$ and $\bar{\mathbf{Y}}$ be the matrices of observables (same as Eq.~\ref{eq:Q_P_def}), each of size $n\times N$, where $n$ is the dimension of the observation function. The standard EDMD estimate involves:
\begin{itemize}
    \item $\bar{\mathbf{Y}}\,\mathbf{Y}^\top$ and $\mathbf{Y}\,\mathbf{Y}^\top$ each require multiplying an $n\times N$ matrix by an $N\times n$ matrix, which costs $\mathcal{O}\left(n^2N\right)$.
    \item Inverting (or pseudo-inverting) $\mathbf{Y}\,\mathbf{Y}^\top$ to get $\mathbf{P}$ typically costs $\mathcal{O}\left(n^3\right)$.
\end{itemize}
Hence, updating $\mathbf{K}$ with every new snapshot in a purely retraining fashion would incur $\mathcal{O}\left(n^2N\right) + \mathcal{O}\left(n^3\right)$ work each time.

\subsection{RLS Update Complexity}\label{sec:rls_update_complexity}

By contrast, the RLS update uses rank-1 updates and the Sherman--Morrison formula to avoid expensive matrix inversions at every step. At time $k\to k+1$, the key operations include:
\begin{itemize}
    \item Updating $\mathbf{P}_{k+1}$ via the Sherman--Morrison formula, which requires a small number of matrix-vector multiplications with matrices and vectors of size $n$, costing $\mathcal{O}\left(n^2\right)$.
    \item Updating $\mathbf{K}_{k+1}$. Like the previous step, this is dominated by a small number of matrix-vector multiplications and outer products at $\mathcal{O}\left(n^2\right)$ each update.
\end{itemize}
Overall, each update has a per-step cost of $\mathcal{O}\left(n^2\right)$.

\section{Convergence Analysis}\label{sec:convergence_analysis}

\subsection{Definitions}\label{sec:definitions}

$\mathcal{S}$ is the state space. The samples in the dataset construct a Markov chain $\{\mathbf{s}_k\}\subset\mathcal{S}$ satisfying $\mathbf{s}_{k+1}=\mathbf{f}\left(\mathbf{s}_k\right)$ for some map $\mathbf{f}:\mathcal{S}\rightarrow\mathcal{S}$. If control inputs are considered, $\mathbf{s}$ contains both the robot state $\mathbf{x}$ and the control input $\mathbf{u}$. In addition, $\boldsymbol{\varphi}:\mathcal{S}\rightarrow\mathbb{R}^{n_{\varphi}}$ is a finite-dimensional observation function applied on $\mathbf{s}$.

The chain $\left\{\mathbf{s}_k\right\}$ admits a stationary distribution (invariant measure) $\mu$ --- if the system starts in the stationary distribution, it remains in that distribution after any number of transitions. Intuitively, we can think of this distribution as describing where the system tends to spend its time. It is an inherent property of the dynamical system and becomes apparent as time approaches infinity. Under the invariant measure $\mu$, define
\begin{equation}\label{eq:Q_def}
    \mathbf{Q}=\mathbb{E}_\mu\left[\boldsymbol{\varphi}\left(\mathbf{f}\left(\mathbf{s}\right)\right)\boldsymbol{\varphi}\left(\mathbf{s}\right)^\top\right],
\end{equation}
\begin{equation}\label{eq:P_def}
    \mathbf{P}=\left(\mathbb{E}_\mu\left[\boldsymbol{\varphi}\left(\mathbf{s}\right)\boldsymbol{\varphi}\left(\mathbf{s}\right)^\top\right]\right)^\dagger.
\end{equation}
The optimal estimate of the Koopman matrix in the mean-square sense in the feature space defined by $\boldsymbol{\varphi}$ is
\begin{equation}
    \mathbf{K}^*=\operatorname*{arg min}_\mathbf{K}~\mathbb{E}_\mu\left[\left\|\boldsymbol{\varphi}\left(\mathbf{f}\left(\mathbf{s}\right)\right)-\mathbf{K}\boldsymbol{\varphi}\left(\mathbf{s}\right)\right\|^2\right].
\end{equation}
And it can be calculated by~\citep{boyd2018}
\begin{equation}
    \mathbf{K}^*=\mathbf{Q}\mathbf{P}.
\end{equation}

To estimate Eq.~\ref{eq:Q_def} and~\ref{eq:P_def} from data, based on EDMD and Eq.~\ref{eq:Q_P_def}, we use
\begin{equation}
    \mathbf{Q}_N=\frac{1}{N}\sum_{k=0}^{N-1}\boldsymbol{\varphi}\left(\mathbf{f}\left(\mathbf{s}_k\right)\right)\boldsymbol{\varphi}\left(\mathbf{s}_k\right)^\top,
\end{equation}
\begin{equation}
    \mathbf{P}_N=\left(\frac{1}{N}\sum_{k=0}^{N-1}\boldsymbol{\varphi}\left(\mathbf{s}_k\right)\boldsymbol{\varphi}\left(\mathbf{s}_k\right)^\top\right)^\dagger.
\end{equation}
where $N$ is the number of data. Then the estimated Koopman matrix calculated by EDMD and RLS is (Appendix~\ref{sec:rls=edmd})
\begin{equation}
    \mathbf{K}_N = \mathbf{Q}_N\mathbf{P}_N.
\end{equation}
Our goal is to show $\mathbf{K}_N$ converges to the optimal estimate $\mathbf{K}^*$.

\subsection{Assumptions}\label{sec:assumptions}

The following assumptions are required in order to guarantee convergence.
\begin{itemize}
    \item $\{\mathbf{s}_k\}$ is a Markov chain. It is irreducible, aperiodic, and positive recurrent w.r.t.\ its invariant measure $\mu$. This implies the chain is ergodic.
    \item The observation function $\boldsymbol{\varphi}\left(\mathbf{s}\right)$ is $\mu$-square-integrable, i.e.,
    \begin{equation}
        \mathbb{E}_\mu\left[\|\boldsymbol{\varphi}\left(\mathbf{s}\right)\|^2\right]=\int_\mathcal{S}\|\boldsymbol{\varphi}\left(\mathbf{s}\right)\|^2d\mu\left(\mathbf{s}\right)<\infty.
    \end{equation}
    \item $\mathbf{P}$ and $\mathbf{P}_N$ are invertible, i.e. they are both full-rank.
\end{itemize}


In our experiments, data are collected via continuous sampling at very short time intervals (10 ms). Consequently, the resulting data trajectories can be approximated as Markov chains, consistent with our state-space model. The experimental results indicate the feasibility of this assumption.

Irreducibility requires that from any sampled state $\mathbf{s}$, there is a nonzero probability of reaching any other region of the state space within a finite number of steps. This condition can be satisfied by the two experimental systems we use because both of them are controllable systems over a bounded domain.

A Markov chain is aperiodic if there is no fixed cycle governing the return of states. For controlled dynamical systems, whether this condition can be satisfied is related to the control policy.

A state $\mathbf{s}$ in a Markov chain is positive recurrent if, starting from $\mathbf{s}$, the chain is guaranteed to return to $\mathbf{s}$ at some future time. For fully actuated systems with bounded control inputs and state space, such as the two used in this study, we can reasonably expect the resulting Markov chains to be positive recurrent.

Regarding the assumption about the rank of $\mathbf{P}$ and $\mathbf{P}_N$, since the components of the observation function form an orthonormal set of basis functions and real mechanical systems exhibit uncertainty and noise, this assumption is always satisfied in practice. In our experiments, we check the rank of $\mathbf{P}_N$ and find that it is always full-rank.

\subsection{Convergence}\label{sec:convergence}

Our assumptions enable us to apply the Strong Law of Large Numbers for Markov chains~\citep{breiman1960} to the sequence $\left\{\mathbf{s}_k\right\}$. The SLLN states that for any integrable function $h:\mathcal{S}\rightarrow\mathbb{R}$, the following statement holds under our assumptions,
\begin{equation}
    \lim_{N\rightarrow\infty}\frac{1}{N}\sum_{k=0}^{N-1}h\left(\mathbf{s}_k\right)=\mathbb{E}_{\mu}\left[h\left(\mathbf{s}\right)\right], \text{ almost surely}.
\end{equation}
Thus for each fixed $i,j\in\{1,...,n_\varphi\}$,
\begin{equation}
    \lim_{N\rightarrow\infty}\frac{1}{N}\sum_{k=0}^{N-1}\varphi_i\left(\mathbf{s}_k\right)\varphi_j\left(\mathbf{s}_k\right)^\top=\mathbb{E}_\mu\left[\varphi_i\left(\mathbf{s}\right)\varphi_j\left(\mathbf{s}\right)^\top\right],
\end{equation}
\begin{equation}
    \lim_{N\rightarrow\infty}\frac{1}{N}\sum_{k=0}^{N-1}\varphi_i\left(\mathbf{s}_k\right)\varphi_j\left(\mathbf{s}_{k+1}\right)^\top=\mathbb{E}_\mu\left[\varphi_i\left(\mathbf{s}\right)\varphi_j\left(\mathbf{f}\left(\mathbf{s}\right)\right)^\top\right],
\end{equation}
almost surely. Hence, we can determine that
\begin{equation}
    \mathbf{Q}_N\xrightarrow[N\rightarrow\infty]{\text{a.s.}}\mathbf{Q},\mathbf{P}_N^{-1}\xrightarrow[N\rightarrow\infty]{\text{a.s.}}\mathbf{P}^{-1}.
\end{equation}
We use $-1$ instead of $\dagger$ here because we have required $\mathbf{P}$ to be full-rank. By standard results on the matrix inverse~\citep{stewart1969},
\begin{equation}
    \mathbf{P}_N\xrightarrow[N\rightarrow\infty]{\text{a.s.}}\mathbf{P}.
\end{equation}
Finally, we can conclude that
\begin{equation}
    \mathbf{K}_N=\mathbf{Q}_N\mathbf{P}_N\xrightarrow[N\rightarrow\infty]{\text{a.s.}}\mathbf{Q}\mathbf{P}=\mathbf{K}^*.
\end{equation}

\section{Details of Experiments}\label{sec:details_of_experiments}

\subsection{Planar Two-Link Arm}\label{sec:planar_two_link_arm}

The simulation environment is a customized MuJoCo environment based on the Gymnasium Reacher-v4 environment~\citep{gymnasium_reacher}, with modifications to the observations, reward function, action space, and time step length.

Instead of observing the sine and cosine of the joint angles, as in Reacher-v4, our customized environment directly observes the joint angles in radians. To enable joint position control, the new reward function is defined as:
\begin{equation}
    \label{eq:reacher_reward} r = -100.0\left\|\mathbf{q}_d-\mathbf{q}\right\| - 10.0\left\|\dot{\mathbf{q}}_d-\dot{\mathbf{q}}\right\| - u_1^2 - u_2^2,
\end{equation}
where $\mathbf{q}_d\in\mathbb{R}^2$ and $\dot{\mathbf{q}}_d\in\mathbb{R}^2$ are the desired joint positions and velocities, $\mathbf{q}\in\mathbb{R}^2$ and $\dot{\mathbf{q}}\in\mathbb{R}^2$ are the actual joint positions and velocities, and $u_1\in\mathbb{R}$ and $u_2\in\mathbb{R}$ are the joint actuator torques. The action space is defined as $u_1, u_2\in[-0.5, 0.5]$ Nm. The time step length is set to 10 ms.

During training, the initial positions of the two joints are initialized by uniformly sampling from $\left[-0.1, 0.1\right]$ rad, and the initial joint velocities are uniformly sampled from $\left[-0.005, 0.005\right]$ rad/s. The target joint positions are uniformly sampled from $\left[-\frac{\pi}{2}, \frac{\pi}{2}\right]$ rad for joint 1 and $\left[-\pi + 0.15, \pi - 0.15\right]$ rad for joint 2. During evaluations, the initial joint state is always set to the first state in the reference trajectory (Eq.~\ref{eq:ref_traj_8}).

\subsubsection{Observation Function}\label{sec:observation_function}

We define the state $\mathbf{x}\in\mathbb{R}^4=\left[q_1~q_2~\dot{q_1}~\dot{q_2}\right]$ and the action $\mathbf{u}=\left[u_1~u_2\right]$. The observation function $\boldsymbol{\phi}\left(\mathbf{x}\right)\in\mathbb{R}^{17}$ we use in RKL and KL is
\begin{equation}\label{eq:reacher_obs}
    \boldsymbol{\phi}\left(\mathbf{x}\right)^\top=
    \begin{bmatrix}
        \mathbf{q}^\top & \dot{\mathbf{q}}^\top & 1 & \text{poly}\left(\mathbf{x}\right)^\top & \text{tri}\left(\mathbf{x}\right)^\top
    \end{bmatrix},
\end{equation}
where
\begin{equation}
    \text{poly}\left(\mathbf{x}\right)^\top=
    \begin{bmatrix}
        q_1^2 & q_1q_2 & q_2^2 & \dot{q_1}^2 & \dot{q_1}\dot{q_2} & \dot{q_2}^2
    \end{bmatrix},
\end{equation}
\begin{equation}
    \text{tri}\left(\mathbf{x}\right)^\top=
    \begin{bmatrix}
        \sin{q_1} & \sin{q_2} & \sin{\left(q_1+q_2\right)} & \cos{q_1} & \cos{q_2} & \cos{\left(q_1+q_2\right)}
    \end{bmatrix}.
\end{equation}

\subsubsection{Reference Trajectory}\label{sec:reference_trajectory}

The reference tip position trajectory is defined by
\begin{equation}\label{eq:ref_traj_8}
    p_x=0.05\sin{\left(\frac{4\pi}{T}t\right)}+0.1, p_y=0.1\cos{\left(\frac{2\pi}{T}t\right)}, t\in\left[0,T\right],
\end{equation}
where T in the period. We choose $T=5$ seconds. The reference joint position trajectory is calculated using inverse kinematics, and the reference joint velocity trajectory is obtained by finite difference.

\subsubsection{KL and RKL Implementation}\label{sec:kl_rkl_implementation}

\begin{figure}[htbp]
    \centering
    \includegraphics[width=\linewidth]{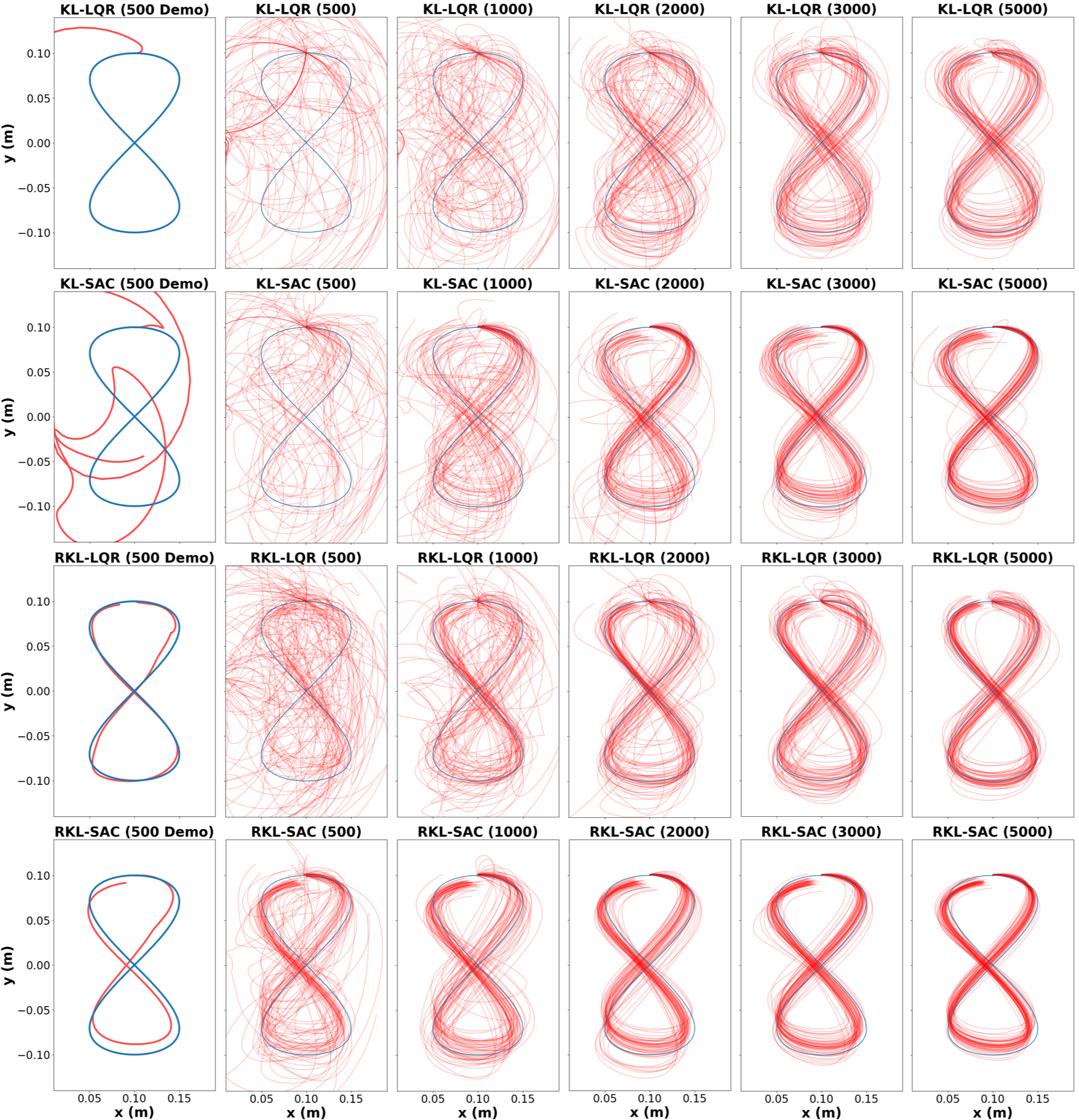}
    \caption{The results of tracking the reference tip position trajectory on the simulated planar two-link arm using KL and RKL with different initial dataset size. Since the generation of the dataset is random, except for the 500-demo dataset, 50 trials are conducted for each plot except for the first column. The red trajectory indicates the actual tip position, and the blue trajectory is the reference. By observing each column, it is obvious that RKL consistently demonstrates superior performance compared to KL. Furthermore, methods based on MPC-SAC outperform those based on LQR.}
    \label{fig:kl_rkl_8_traj}
\end{figure}

We implement KL and RKL in Python, using a single thread. Every 10 ms, we compute the optimal action, update the model (if using RKL), and simulate the system one step forward. We test six different initial dataset sizes (500 Demo, 500, 1,000, 2,000, 3,000, and 5,000 data points) and two types of MPC solver (MPC-SAC and LQR) for both KL and RKL. The 500 Demo initial dataset is generated by using a PD controller to control the joints, and other initial datasets are generated by applying random actions uniformly sampled from the action space. For each of the resulting 24 policies, we perform 50 trials. The results are shown in Table~\ref{fig:reacher_data_table} in the main paper and visualized in Fig.~\ref{fig:kl_rkl_8_traj}.

\subsubsection{NN-MPPI Training}\label{sec:nn-mppi_training}

\begin{figure}[htbp]
    \centering
    \includegraphics[width=0.6\linewidth]{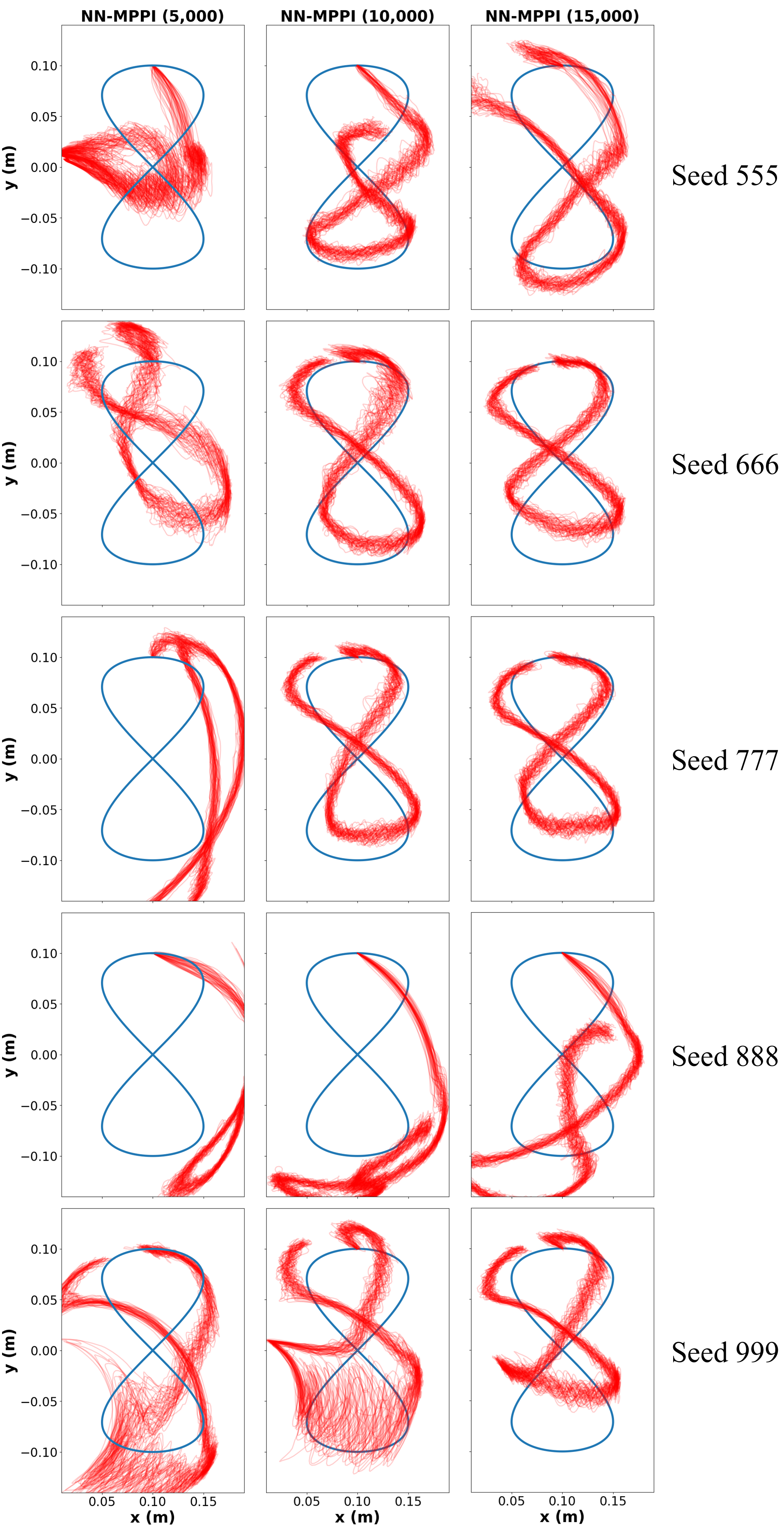}
    \caption{The results of tracking the reference tip position trajectory on the simulated planar two-link arm using policies trained by NN-MPPI. Each row corresponds to a distinct, fixed random seed, while each column corresponds to a different training steps. The red trajectory indicates the actual tip position, and the blue trajectory is the reference. Trajectories generated by NN-MPPI exhibit significant jitter, and the performance consistency across different random seeds is low.}
    \label{fig:mppi_8_traj}
\end{figure}

We use the NN-MPPI implementation from MaxDiffRL~\cite{berrueta2024}. We train 15 policies using five random seeds (555, 666, 777, 888, and 999) and three different total numbers of steps (5,000, 10,000, and 15,000). For each trained policy, 50 trials are performed. The average results across all 5 random seeds are shown in Table~\ref{fig:reacher_data_table} in the main paper for each "Dataset size" and visualized in Fig.~\ref{fig:mppi_8_traj}. The parameters for training and the MPPI controller are shown in Table~\ref{tab:tip_track_nn_mppi}.

\subsubsection{RL-SAC Training}\label{sec:rl-sac_training}

\begin{figure}[htbp]
    \centering
    \includegraphics[width=0.75\linewidth]{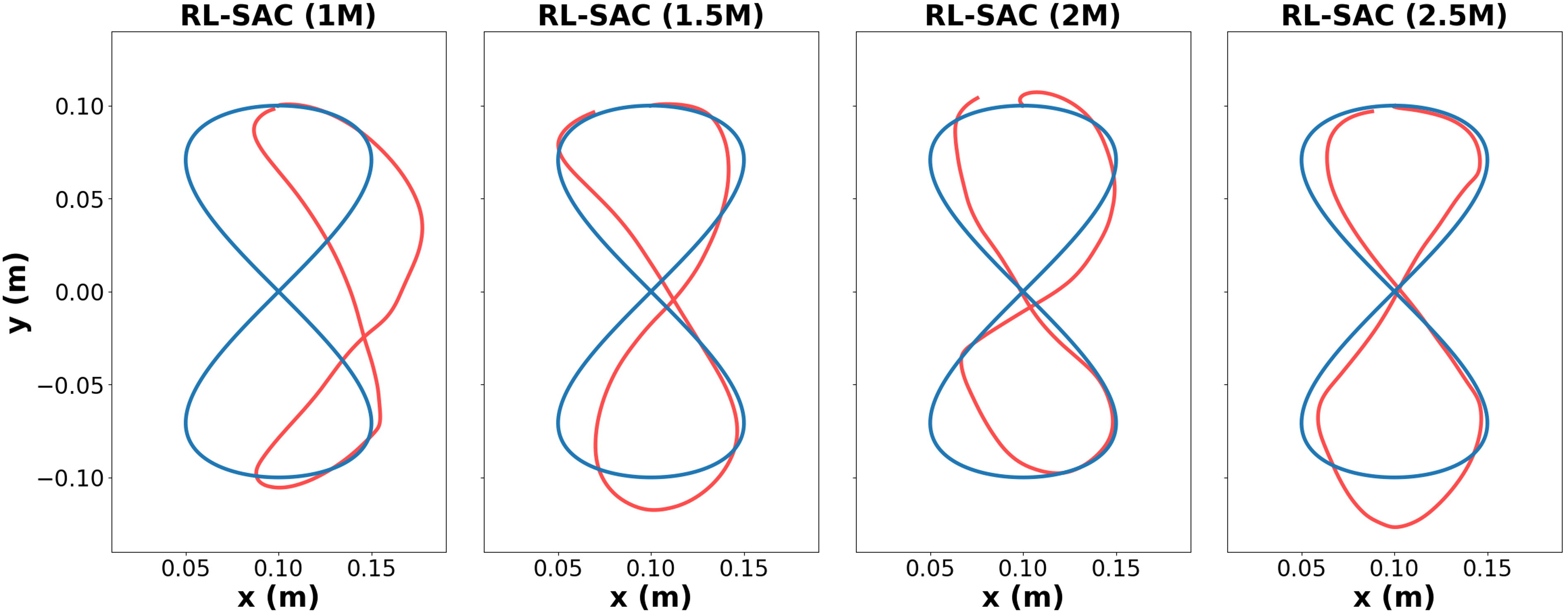}
    \caption{The results of tracking the reference tip position trajectory on the simulated planar two-link arm using policies trained by RL-SAC, with training steps 1 million to 2.5 million. The red trajectory indicates the actual tip position, and the blue trajectory is the reference. Despite the ability of RL-SAC to achieve adequate control, its sample efficiency is notably low.}
    \label{fig:sac_8_traj}
\end{figure}

We use the RL-SAC implementation from stable-baselines3~\cite{stable-baselines3}. Training is performed on 32 independently and randomly initialized environments with different seeds using SubprocVecEnv. The evaluation results are shown in Table~\ref{fig:reacher_data_table} in the main paper and visualized in Fig.~\ref{fig:sac_8_traj}. The training parameters are detailed in Table~\ref{tab:tip_track_sac}.

\subsubsection{REDQ Training}\label{sec:redq_training}

\begin{figure}[htbp]
    \centering
    \includegraphics[width=0.55\linewidth]{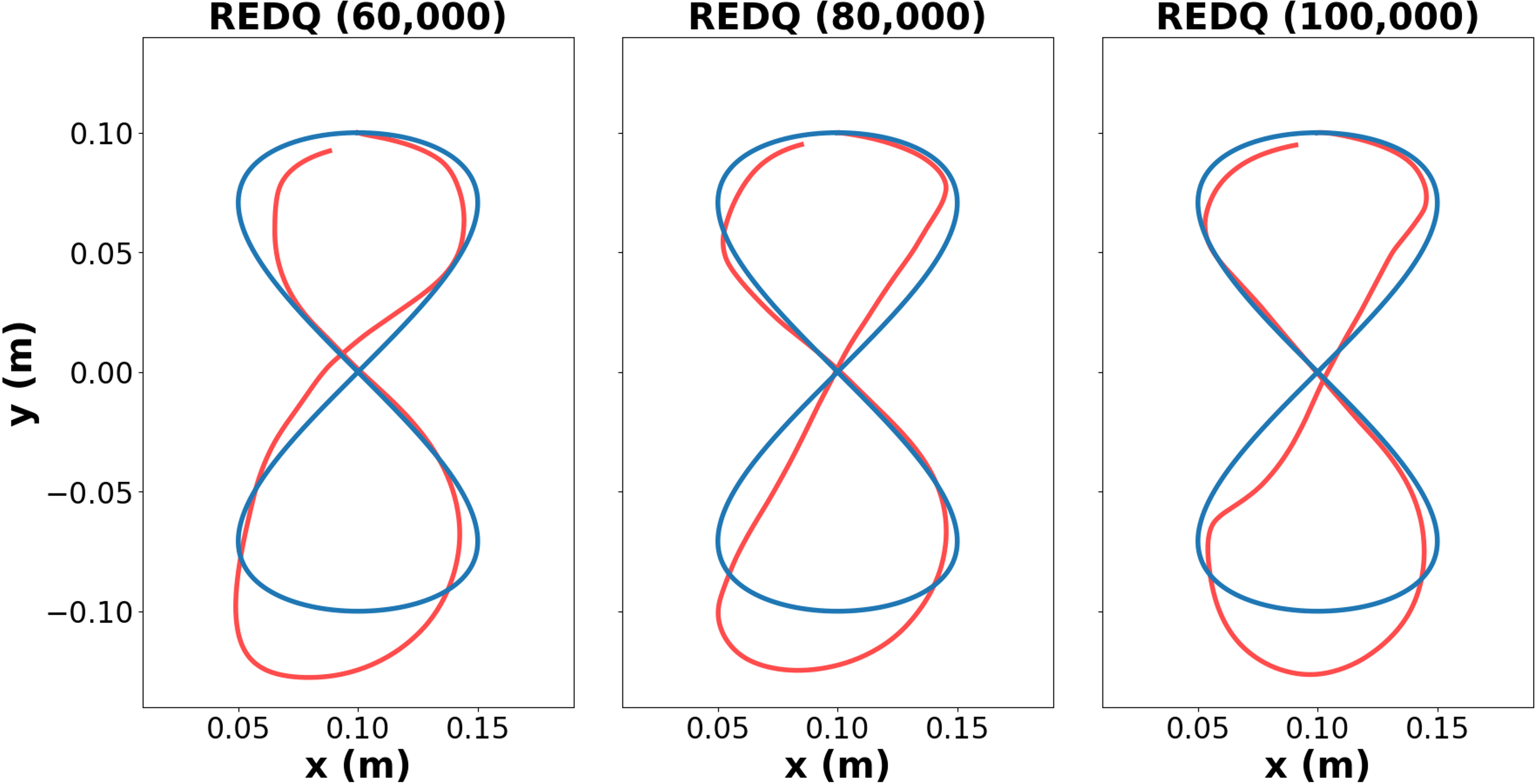}
    \caption{The results of tracking the reference tip position trajectory on the simulated planar two-link arm using policies trained by REDQ, with training steps 60K, 80K, and 100K. The red trajectory indicates the actual tip position, and the blue trajectory is the reference. As~\citep{chen2021} states, REDQ does have high sample efficiency compared to other model-free RL methods, but RKL still significantly outperforms it.}
    \label{fig:redq_8_traj}
\end{figure}

We use the open-source code provided by the authors of REDQ~\cite{chen2021}. The evaluation results are shown in Table~\ref{fig:reacher_data_table} in the main paper and in Fig.~\ref{fig:redq_8_traj}. The training parameters are detailed in Table~\ref{tab:tip_track_redq}.

\begin{table}[h]
    \centering
    \begin{minipage}{0.49\textwidth}
        \centering
        \begin{tabular}{|c|c|}
            \hline
            Horizon & 160 ms \\ \hline
            LQR $\mathbf{q}$ weight & 200.0 \\ \hline
            LQR $\mathbf{\dot{q}}$ weight & 30.0 \\ \hline
            LQR observations weight & 1.0 \\ \hline
            LQR $\mathbf{u}$ weights & 0.001 \\ \hline
            LQR terminal weights & 0.0 \\ \hline
            MPC-SAC $\mathbf{u}$ weights & 1e10 \\ \hline
        \end{tabular}
        \vspace{0.5cm}
        \caption{KL-SAC and RKL-SAC on the Planar Two-link Arm}
        \label{tab:tip_track_rkl_sac}
    \end{minipage}    
    \hfill
    \begin{minipage}{0.49\textwidth}
        \centering
        \begin{tabular}{|c|c|}
            \hline
            Horizon & 500 ms \\ \hline
            LQR $\mathbf{p}$ weight & 200.0 \\ \hline
            LQR $\mathbf{\dot{q}}$ weight & 5.0 \\ \hline
            LQR observations weight & 1.0 \\ \hline
            LQR $\mathbf{u}$ weights & 0.001 \\ \hline
            LQR terminal weights & 0.0 \\ \hline
        \end{tabular}
        \vspace{0.5cm}
        \caption{KL-LQR and RKL-LQR Parameters on the Planar Two-link Arm}
        \label{tab:tip_track_rkl_lqr}
    \end{minipage}    
    \begin{minipage}{0.49\textwidth}
        \centering
        \begin{tabular}{|c|c|}
            \hline
            Horizon & 300 ms \\ \hline
            Number of samples & 2000 \\ \hline
            Temperature & 0.5 \\ \hline
            Gaussian noise std dev & 0.5 \\ \hline
            Model learning rate & 3e-4 \\ \hline
            Model training iterations & 6 \\ \hline
            Model training batch size & 128 \\ \hline
        \end{tabular}
        \vspace{0.5cm}
        \caption{NN-MPPI Parameters on the Planar Two-link Arm}
        \label{tab:tip_track_nn_mppi}
    \end{minipage}    
    \hfill
    \begin{minipage}{0.49\textwidth}
        \centering
        \begin{tabular}{|c|c|}
            \hline
            Batch size & 256 \\ \hline
            Learning rate & 3e-4 \\ \hline
            Buffer size & 1e6 \\ \hline
            Learning start & 1e4 \\ \hline
            Discount factor & 0.99 \\ \hline
            Training frequency & 1 \\ \hline
            Target network update rate & 0.005 \\ \hline
        \end{tabular}
        \vspace{0.5cm}
        \caption{RL-SAC Training Parameters on the Planar Two-link Arm}
        \label{tab:tip_track_sac}
    \end{minipage}
    \begin{minipage}{0.49\textwidth}
        \centering
        \begin{tabular}{|c|c|}
            \hline
            Batch size & 256 \\ \hline
            Learning rate & 3e-4 \\ \hline
            Learning start & 2000 \\ \hline
            Buffer size & 1e6 \\ \hline
            Discount factor & 0.99 \\ \hline
            Polyak averaging factor & 0.995 \\ \hline
            UTD ratio & 20 \\ \hline
            \#Q-functios & 10 \\ \hline
            \#Q-values when forming the target & 2 \\ \hline
            Q-targets mode & min \\ \hline
            Policy update delay & 20 \\ \hline            
        \end{tabular}
        \vspace{0.5cm}
        \caption{REDQ Training Parameters on the Planar Two-link Arm}
        \label{tab:tip_track_redq}
    \end{minipage}
\end{table}

\subsection{Soft Stewart Platform}\label{sec:soft_stewart_platform}

\begin{figure}[H]
    \centering
    \includegraphics[width=0.3\textwidth]{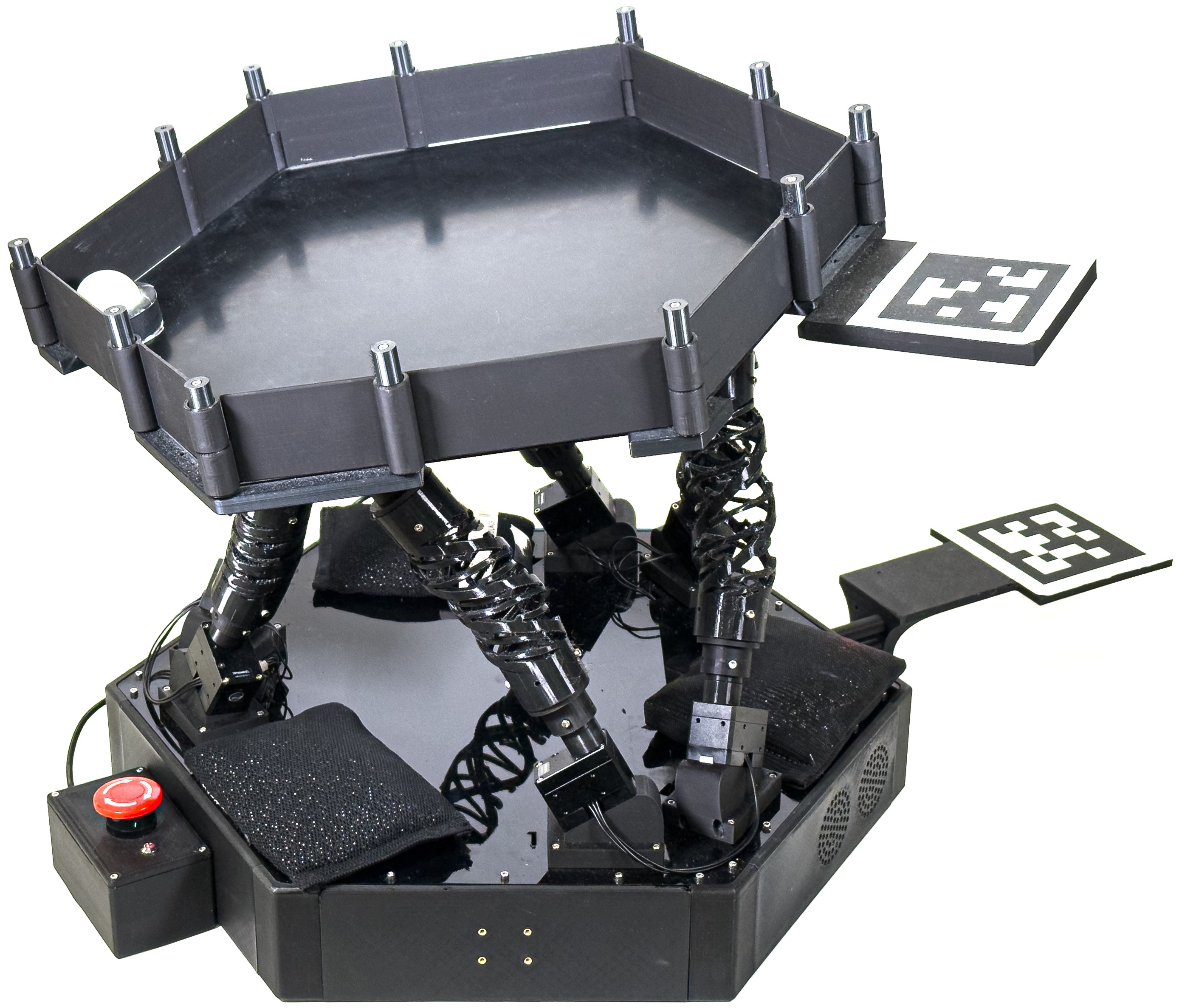}
    \caption{The Soft Stewart Platform}
    \label{fig:soft_stewart_platform}
\end{figure}
\vspace{-10pt}

The Soft Stewart Platform (SSP) is a six-DoF parallel manipulator inspired by the traditional Stewart platform design (Fig.~\ref{fig:soft_stewart_platform}). It is actuated by six Handed Shearing Auxetics (HSAs)~\citep{kim2024}, which are manufactured using fused deposition modeling (FDM) with TPU 95A filament on a Bambu X1‑Carbon printer. Each actuator is anchored to the top of the platform and driven at the opposite end by a DYNAMIXEL XM430‑W350‑T servo motor. Unlike a conventional Stewart platform, this design eliminates joints at the ends of the struts. Instead, the compliance of the HSAs provides the  kinematic behavior typically achieved with such joints. The platform’s top surface is a hexagonal piece of acrylic, with each side measuring 24 cm, sanded to reduce friction. Surrounding walls and edges, made from 3D‑printed polylactic acid (PLA), not only prevent the puck from falling off but also allow intermittent mechanical contact.

State information is obtained using two AprilTags mounted on the top and bottom of the platform. A Logitech BRIO webcam positioned above the setup tracks these tags while simultaneously detecting the balancing puck through a Hough circles algorithm. Using ray-tracing techniques, the system calculates the puck’s position relative to the platform's center. The combined state of the platform-puck system is computed at approximately 60 Hz, with motor commands issued at 50 Hz.

\subsubsection{NN-MPPI Implementation}\label{sec:nn-mppi_implementation}

Model-predictive path integral control (MPPI) is a model predictive control (MPC) algorithm \cite{williams_aggressive_2016}, and NN-MPPI refers to the model-based reinforcement learning framework that uses MPPI alongside neural networks as dynamics models \cite{williams2017}. This was implemented and benchmarked on the Soft Stewart Platform in~\citep{avtges2025}. This approach uses an approximate geometric model of the SSP as if it were rigid, and learns residuals over this model to accomodate the platform's nonlinear actuators. The observation space is defined as $\mathbf{o}\in \mathbb{R}^{9} = [x, y, \dot{x}, \dot{y}, \phi, \theta, \psi, X, Y]$. This vector contains $[x, y, \dot{x}, \dot{y}]$ the position and velocity of the puck, $[\phi, \theta, \psi]$ the Euler angles of the top platform surface relative to the base, and $[X, Y]$ the coordinates of the balancing setpoint. Actions are computed at 15 Hz.

\subsubsection{KL and RKL Implementation}\label{sec:kl_rkl_implementation}

\textit{\textbf{State feedback}}: To model the system, we define the state $\mathbf{x}\in\mathbb{R}^4=\left[p_x,p_y,v_x,v_y\right]$, which includes the position and velocity of the puck, and the control input $\mathbf{u}\in\mathbb{R}^6$, representing the angles of the six servos. The puck’s current position is determined using a computer vision algorithm, and a Moving Average Filter (MAF) with a window size of approximately 65 ms is applied to the raw position feedback. The velocity is then estimated from the filtered positions using finite differences.


\textit{\textbf{Initial dataset}}: The initial dataset is collected by a human operator using a SpaceMouse Compact and an inverse kinematic approximation that models the platform as if it was rigid. The purpose of this dataset is to create a reasonable initial model, thereby mitigating numerical issues, so no specific control objectives are provided during data collection. The same initial dataset is used for all experiments, and the distribution of the puck’s positions in this dataset is depicted in Fig.~\ref{fig:initial_dataset_position} and~\ref{fig:initial_dataset_velocity}.

Collecting raw state and control input data exhibits fluctuations and occasional missing values. To address this, linear interpolation is applied to align the state and input data to a consistent frequency. The processed initial dataset has a time step length of 10 ms. Subsequently, EDMD is performed on this dataset to obtain the initial Koopman model.

\textit{\textbf{C++ implementation}}: The controller is implemented in C++ using a multi-threaded architecture. Four threads are employed:
\begin{itemize}
    \item \textbf{Feedback thread} receives feedback on the puck’s states.
    \item \textbf{Command thread} sends commands to the servos.
    \item \textbf{Control thread} solves the optimal control problem.
    \item \textbf{Update thread} updates the Koopman model with the latest data.
\end{itemize}

All threads run at a frequency of 50 Hz, constrained by the raw position feedback rate. However, in our experiments, MPC-SAC typically solves the control problem within 15 ms, and the model update is completed within 10 ms. In addition, an MAF is applied to the raw solutions before being sent to the servos.

\subsubsection{KL and RKL Observation Functions}\label{sec:kl_rkl_observation_functions}

We use two types of observation functions in our experiments on the Soft Stewart Platform: a 28-dimensional polynomial basis function up to the third degree and a 117-dimensional Gaussian Radial Basis Function (RBF).

The 28-dimensional polynomial basis function is defined as
\begin{equation}\label{eq:poly_obs}
    \boldsymbol{\phi}\left(\mathbf{x}\right)^\top=
    \begin{bmatrix}
        \mathbf{x}^\top&x_1^2&x_1x_2&\cdots&x_1x_2x_3
    \end{bmatrix},
\end{equation}
while the 117-dimensional Gaussian RBF is expressed as
\begin{equation}\label{eq:gaussian_rbf}
    \boldsymbol{\phi}\left(\mathbf{x}\right)^\top=
    \begin{bmatrix}
        \mathbf{x}^\top&\phi_1\left(\mathbf{x}\right)&\phi_2\left(\mathbf{x}\right)&\cdots&\phi_{117}\left(\mathbf{x}\right)
    \end{bmatrix},
\end{equation}
where each $\phi_i\left(\mathbf{x}\right)$ is defined by
\begin{equation}
    \phi_i\left(\mathbf{x}\right)=\exp\left(-\epsilon\left\|\mathbf{x}-\mathbf{c}_i\right\|_2^2\right).
\end{equation}

To define the centers $\mathbf{c}_i\in\mathbb{R}^4$, we follow the same methodology used to generate the goal points for the experiments. For the position component of $\mathbf{c}_i$, 117 uniformly distributed points are used, with a distance of 3.5 cm between each point and its six adjacent points. The sampling area for these points is slightly larger than the platform to ensure that positions near the boundaries can excite a sufficient number of RBFs, similar to positions within the platform. For the velocity component, $k$-means++ clustering is performed on the initial dataset to identify suitable velocity centers.


\subsubsection{RL-SAC Training}\label{sec:rl-sac_training}

We choose Soft Actor-Critic (RL-SAC)~\citep{haarnoja2018} as a benchmark for learned control. RL-SAC is an off-policy model-free maximum entropy deep reinforcement learning framework and is widely used for continuous control tasks. We train models in a single deployment---that is, non-episodically without environmental resets. We allow RL-SAC to train for 100,000 environment steps at a frequency of 10 Hz. To increase learning stability, we use a curriculum learning procedure that uniformly samples a new balance setpoint every 20 seconds, within a radius of the platform center. This radius is initialized to 3 cm and over the first half of the training procedure, it linearly expands to cover the entire platform. We retain the same input-output relationships as KL and RKL, with the addition of the current setpoint coordinates in the state vector $\mathbf{x}$. Five random seeds of RL-SAC are trained and evaluated. RL-SAC training does not have access to the initial dataset in Sec.~\ref{sec:initial_dataset}.

\subsubsection{Parameters}\label{sec:parameters}

The parameters for puck balancing using KL and RKL with the 28-dimensional polynomial basis function are shown in Table~\ref{tab:puck_balancing_poly}. The parameters for puck balancing using KL and RKL with the 117-dimensional Gaussian RBF are shown in Table~\ref{tab:puck_balancing_rbf}. The parameters for trajectory tracking using KL and RKL with the 28-dimensional polynomial basis function are shown in Table~\ref{tab:traj_track_poly}. The parameters for RL-SAC training are shown in Table~\ref{tab:sac_params}. The parameters for NN-MPPI training are shown in Table~\ref{tab:mppi_params}.

\begin{table}[h]
    \centering
    \begin{minipage}{0.49\textwidth}
        \centering
        \begin{tabular}{|c|c|}
            \hline
            Horizon & 160 ms \\ \hline
            LQR $\mathbf{p}$ weight & 17.0 \\ \hline
            LQR $\mathbf{v}$ weight & 1.0 \\ \hline
            LQR observations weight & 0.0 \\ \hline
            LQR $\mathbf{u}$ weights & 1.5 \\ \hline
            LQR terminal weights & 0.0 \\ \hline
            MPC-SAC $\mathbf{u}$ weights & 0.002 \\ \hline
            $\mathbf{u}$ MAF window size & 120 ms \\ \hline
        \end{tabular}
        \vspace{0.5cm}
        \caption{Puck Balancing using Polynomial Observables}
        \label{tab:puck_balancing_poly}
    \end{minipage}
    \hfill
    \begin{minipage}{0.49\textwidth}
        \centering
        \begin{tabular}{|c|c|}
            \hline
            Horizon & 160 ms \\ \hline
            LQR $\mathbf{p}$ weight & 17.0 \\ \hline
            LQR $\mathbf{v}$ weight & 1.0 \\ \hline
            LQR observations weight & 0.0 \\ \hline
            LQR $\mathbf{u}$ weights & 0.001 \\ \hline
            LQR terminal weights & 0.0 \\ \hline
            MPC-SAC $\mathbf{u}$ weights & 4.25e-5 \\ \hline
            $\mathbf{u}$ MAF window size & 120 ms \\ \hline
        \end{tabular}
        \vspace{0.5cm}
        \caption{Puck Balancing using Gaussain RBF Observables}
        \label{tab:puck_balancing_rbf}
    \end{minipage}
    \begin{minipage}{0.49\textwidth}
        \centering
        \begin{tabular}{|c|c|}
            \hline
            Horizon & 200 ms \\ \hline
            LQR $\mathbf{p}$ weight & 17.0 \\ \hline
            LQR $\mathbf{v}$ weight & 1.5 \\ \hline
            LQR observations weight & 0.0 \\ \hline
            LQR $\mathbf{u}$ weights & 1.5 \\ \hline
            LQR terminal weights & 0.0 \\ \hline
            MPC-SAC $\mathbf{u}$ weights & 0.0015 \\ \hline
            $\mathbf{u}$ MAF window size & 120 ms \\ \hline
        \end{tabular}
        \vspace{0.5cm}
        \caption{Puck Trajectory Tracking using Polynomial Observables}
        \label{tab:traj_track_poly}
    \end{minipage}
    \hfill
    \begin{minipage}{0.49\textwidth}
        \centering
        \begin{tabular}{|c|c|}
            \hline
            State Dim. & 6 \\ \hline
            Action Dim. & 6 \\ \hline
            Learning Rate & 0.0003 \\ \hline
            Batch Size & 128 \\ \hline
            Policy Network Dim. & [256] \\ \hline
            Discount & 0.99 \\ \hline
            Smoothing Coeff. & 0.005 \\ \hline
            Reward Scale & 0.1 \\ \hline
        \end{tabular}
        \vspace{0.5cm}
        \caption{RL-SAC Parameters}
        \label{tab:sac_params}
    \end{minipage}
    \begin{minipage}{0.49\textwidth}
        \centering
        \begin{tabular}{|c|c|}
            \hline
            State Dim. & 9 \\ \hline
            Action Dim. & 2 \\ \hline
            Horizon & 20 steps \\ \hline
            Learning Rate & 0.0003 \\ \hline
            Batch Size & 128 \\ \hline
            Network Dim. & [200, 200] \\ \hline
            Discount & 0.95 \\ \hline
            Samples & 4096 \\ \hline
            Temperature $\lambda$ & 0.1 \\ \hline
        \end{tabular}
        \vspace{0.5cm}
        \caption{NN-MPPI Parameters}
        \label{tab:mppi_params}
    \end{minipage}
\end{table}

\section{Initial Dataset for SSP Experiments}\label{sec:initial_dataset}

\begin{figure}[htbp]
    \centering
    \begin{subfigure}[b]{0.216\textwidth}
        \centering
        \includegraphics[width=\textwidth]{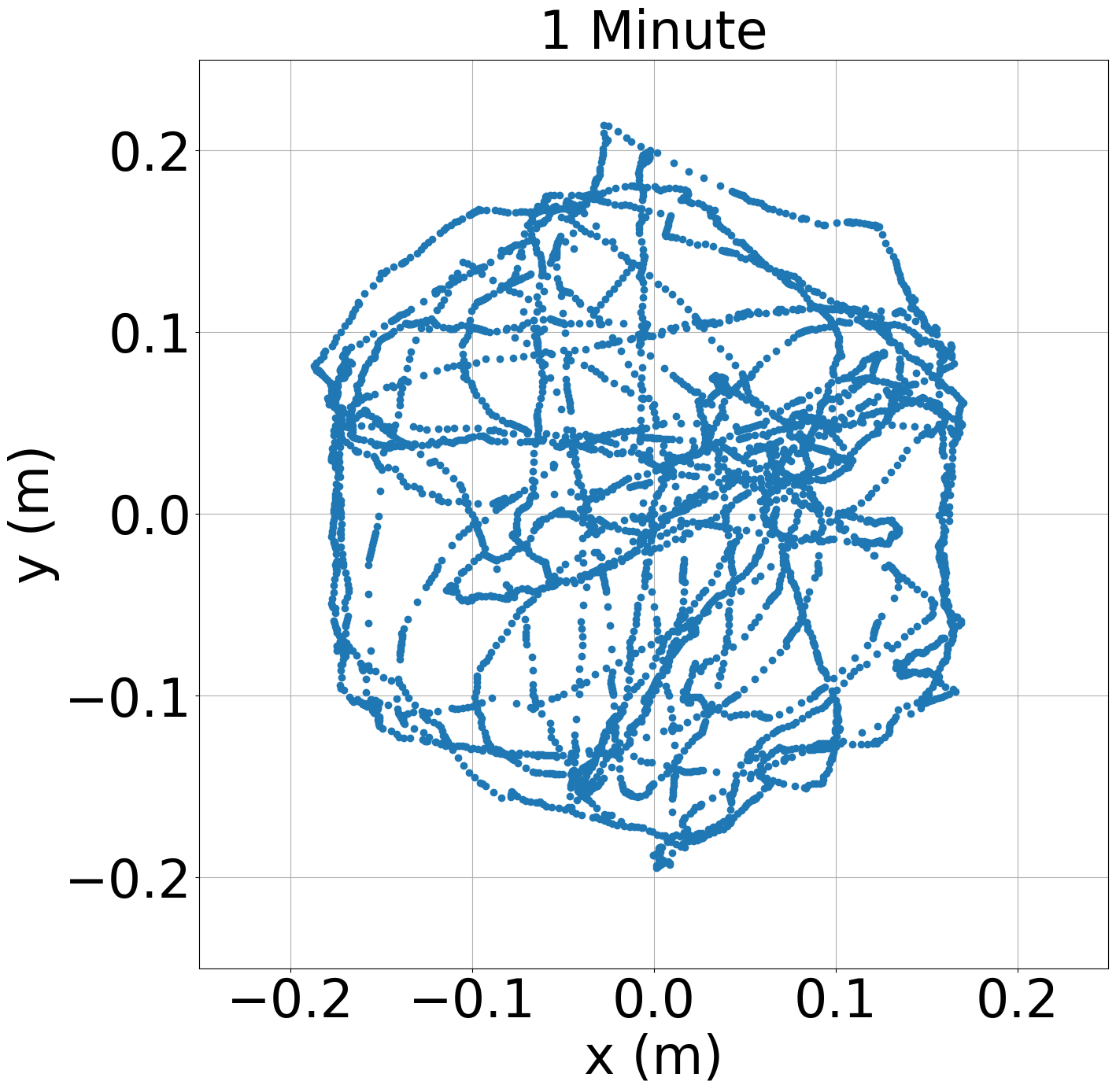}
    \end{subfigure}
    \begin{subfigure}[b]{0.18\textwidth}
        \centering
        \includegraphics[width=\textwidth]{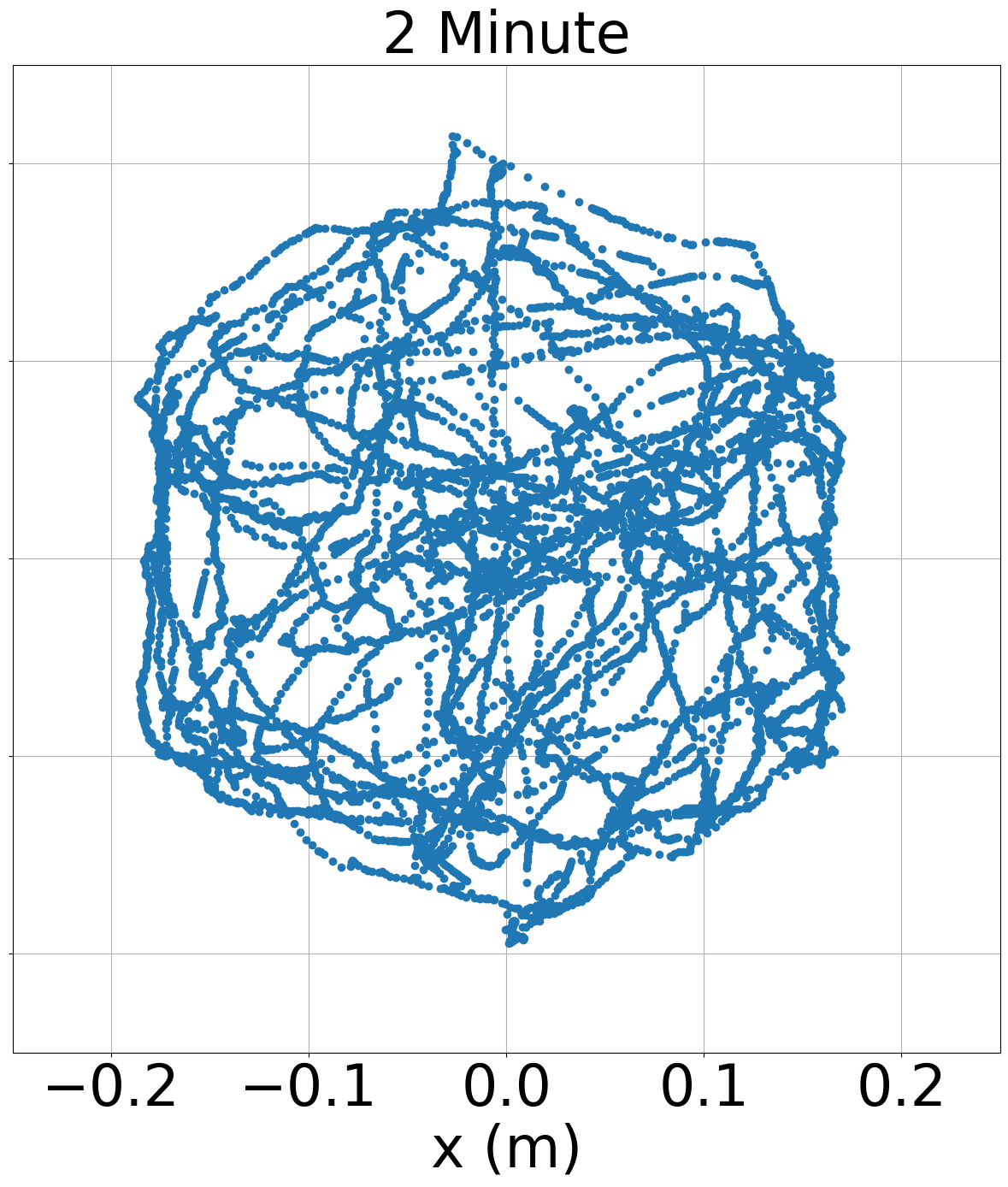}
    \end{subfigure}
    \begin{subfigure}[b]{0.18\textwidth}
        \centering
        \includegraphics[width=\textwidth]{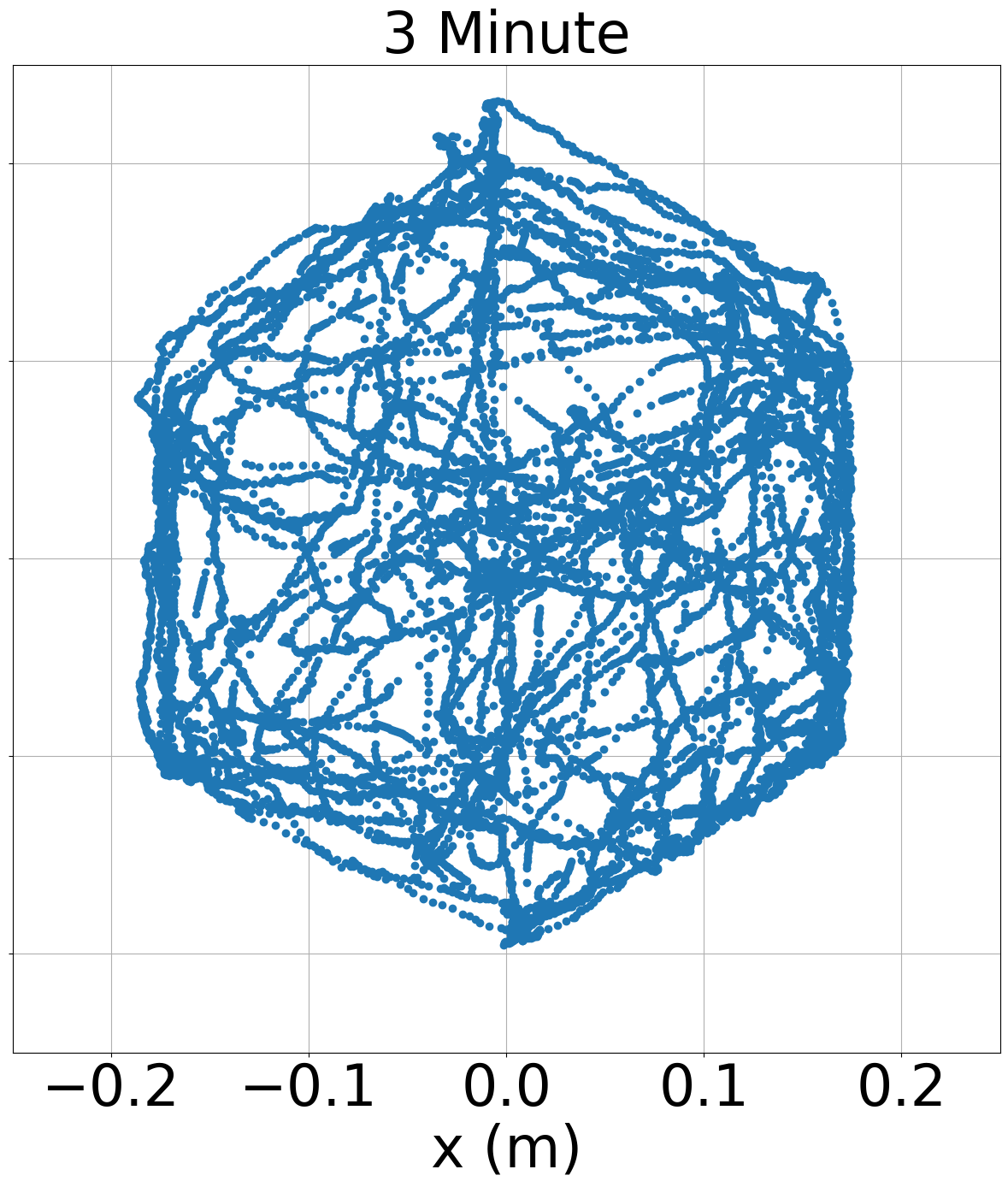}
    \end{subfigure}
    \begin{subfigure}[b]{0.18\textwidth}
        \centering
        \includegraphics[width=\textwidth]{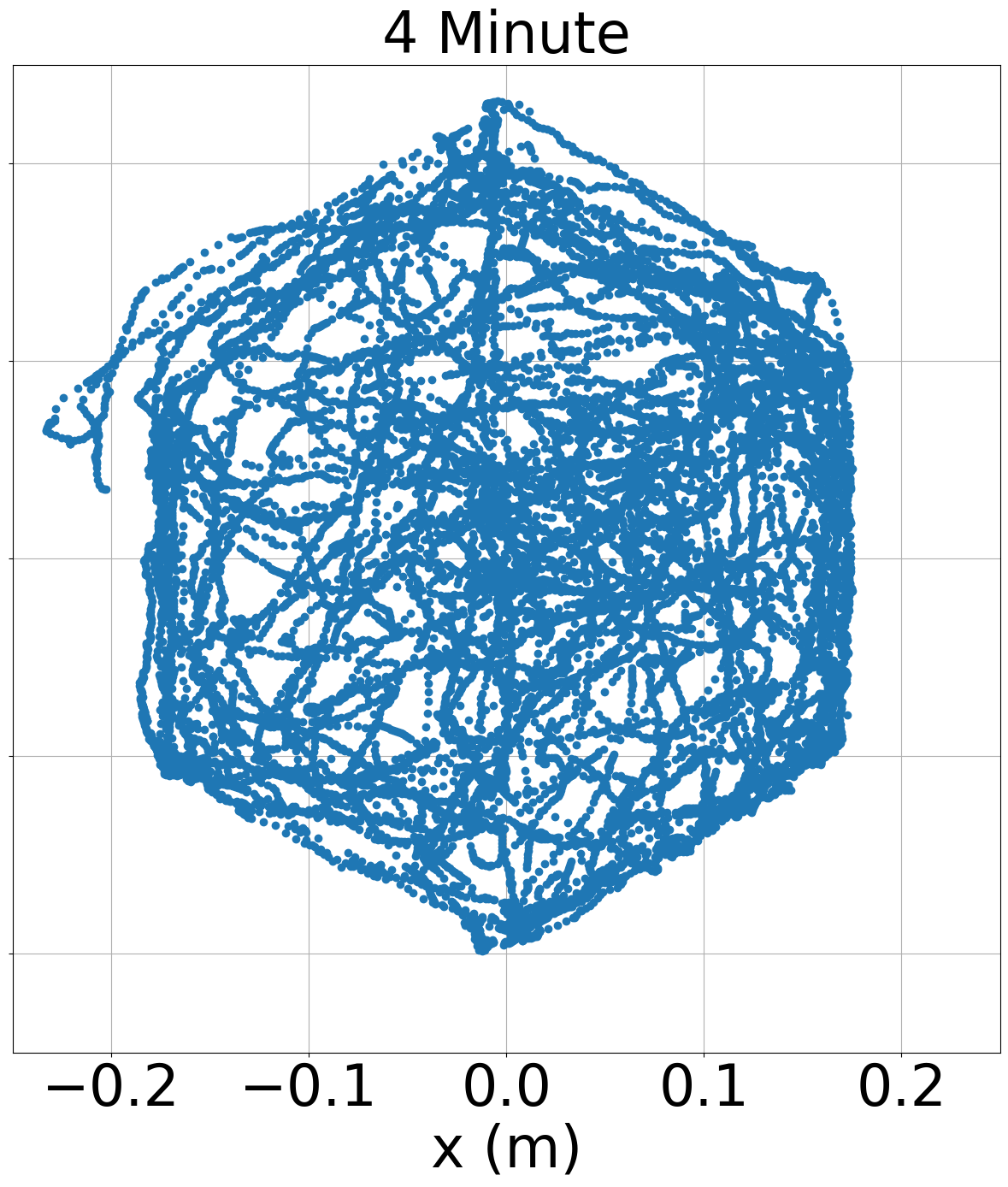}
    \end{subfigure}
    \begin{subfigure}[b]{0.18\textwidth}
        \centering
        \includegraphics[width=\textwidth]{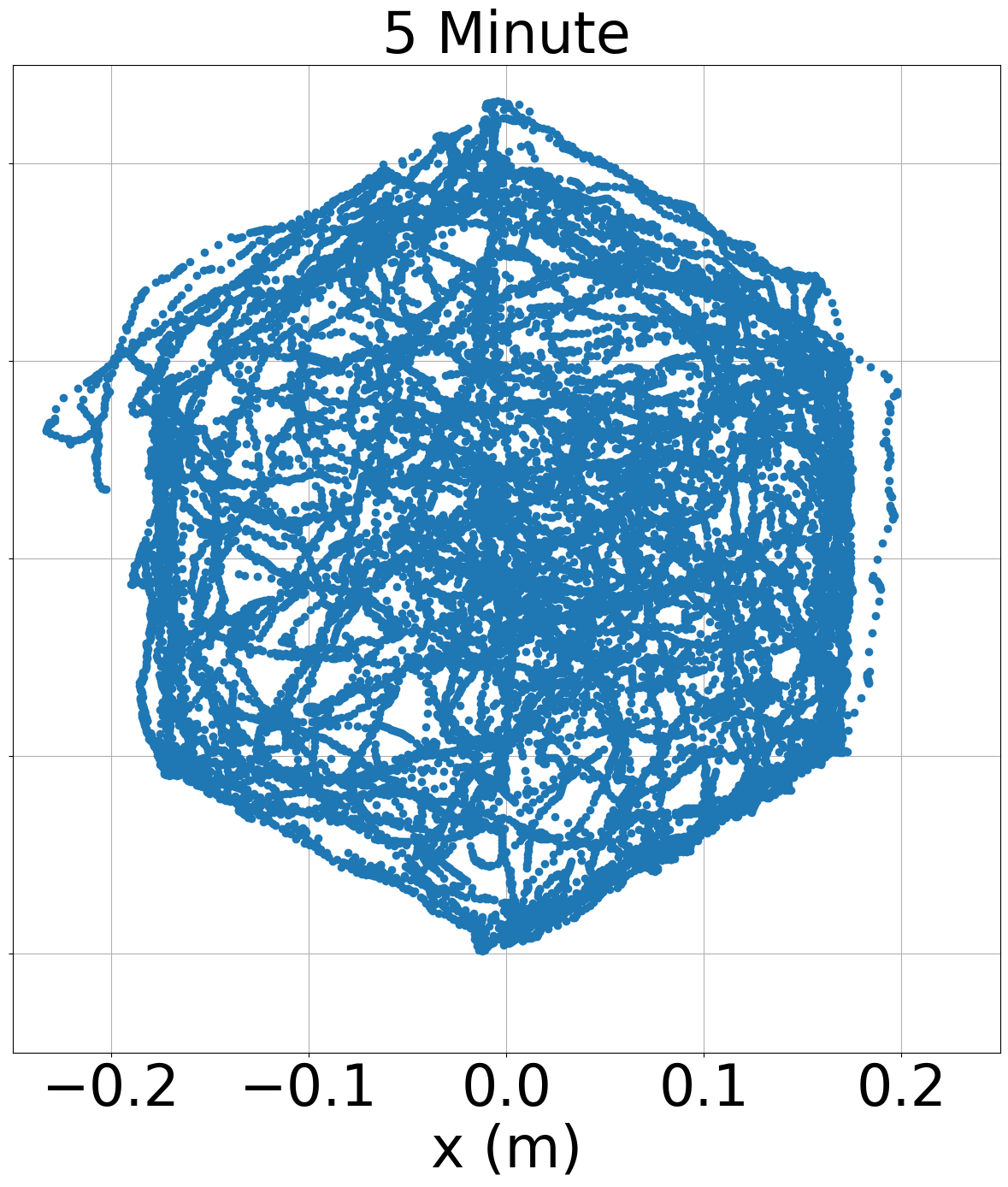}
    \end{subfigure}
    \caption{The distribution of the puck's positions (m) in the initial dataset.}
    \label{fig:initial_dataset_position}
\end{figure}

\begin{figure}[htbp]
    \centering
    \begin{subfigure}[b]{0.215\textwidth}
        \centering
        \includegraphics[width=\textwidth]{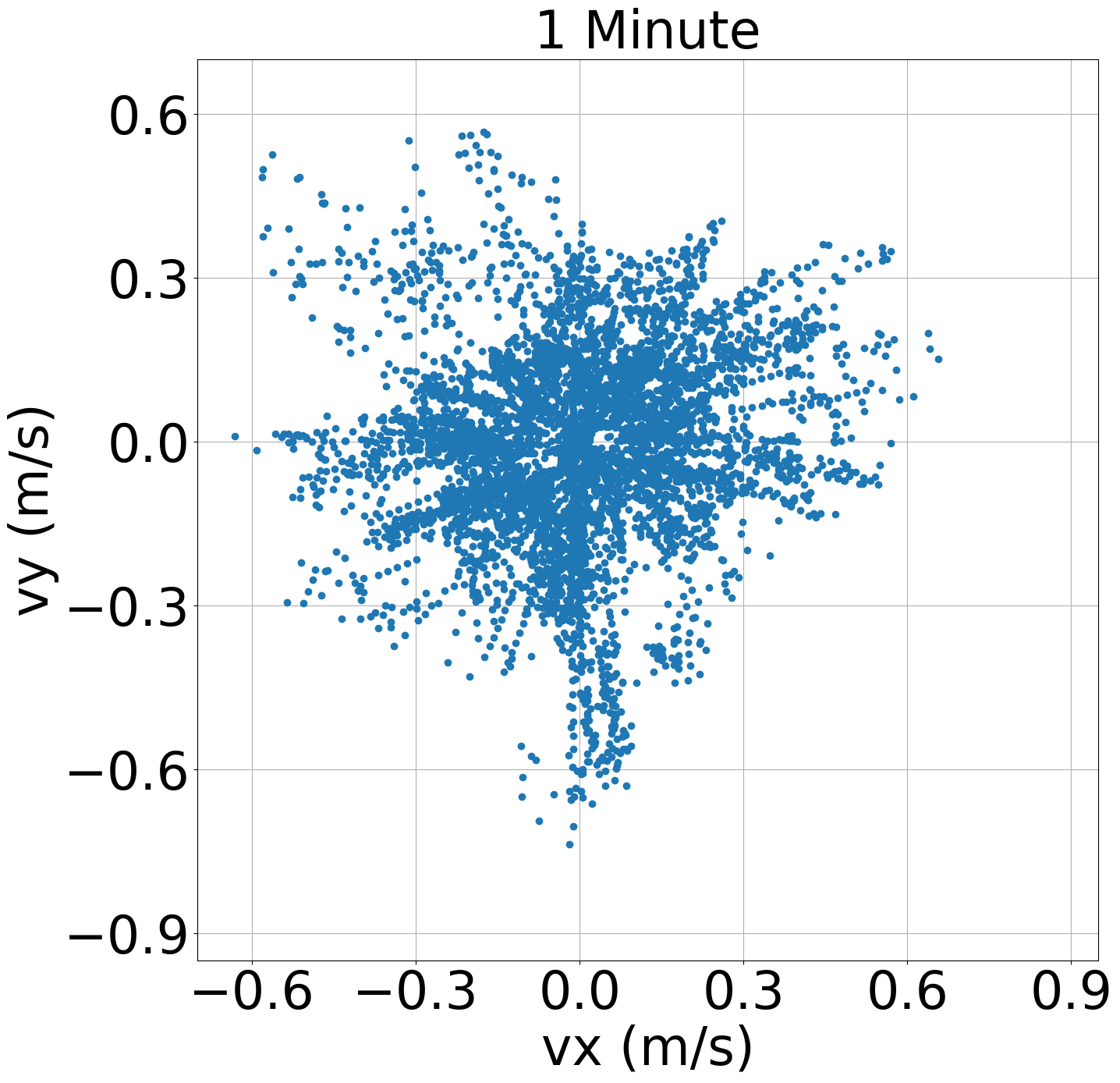}
    \end{subfigure}
    \begin{subfigure}[b]{0.18\textwidth}
        \centering
        \includegraphics[width=\textwidth]{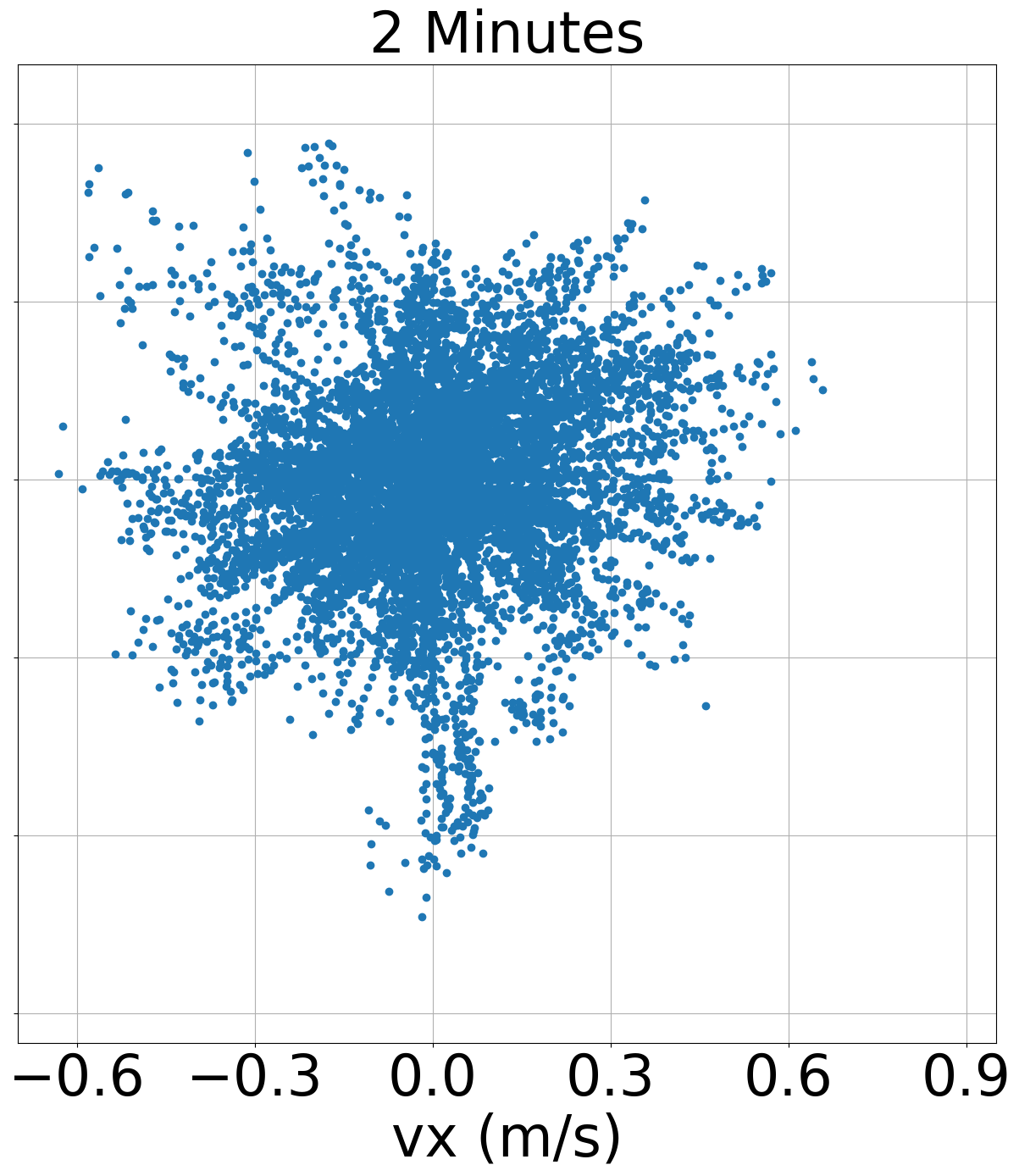}
    \end{subfigure}
    \begin{subfigure}[b]{0.18\textwidth}
        \centering
        \includegraphics[width=\textwidth]{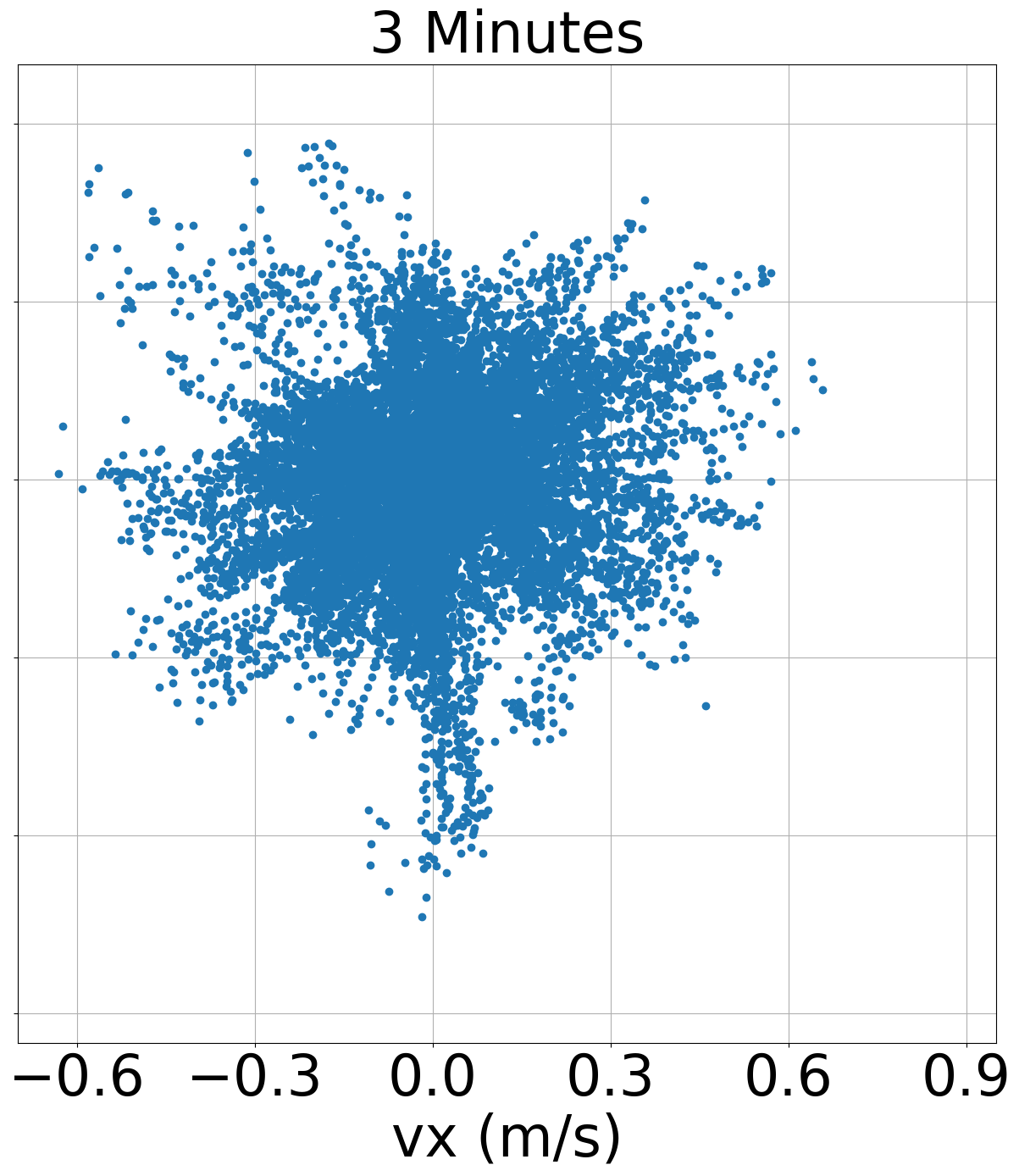}
    \end{subfigure}
    \begin{subfigure}[b]{0.18\textwidth}
        \centering
        \includegraphics[width=\textwidth]{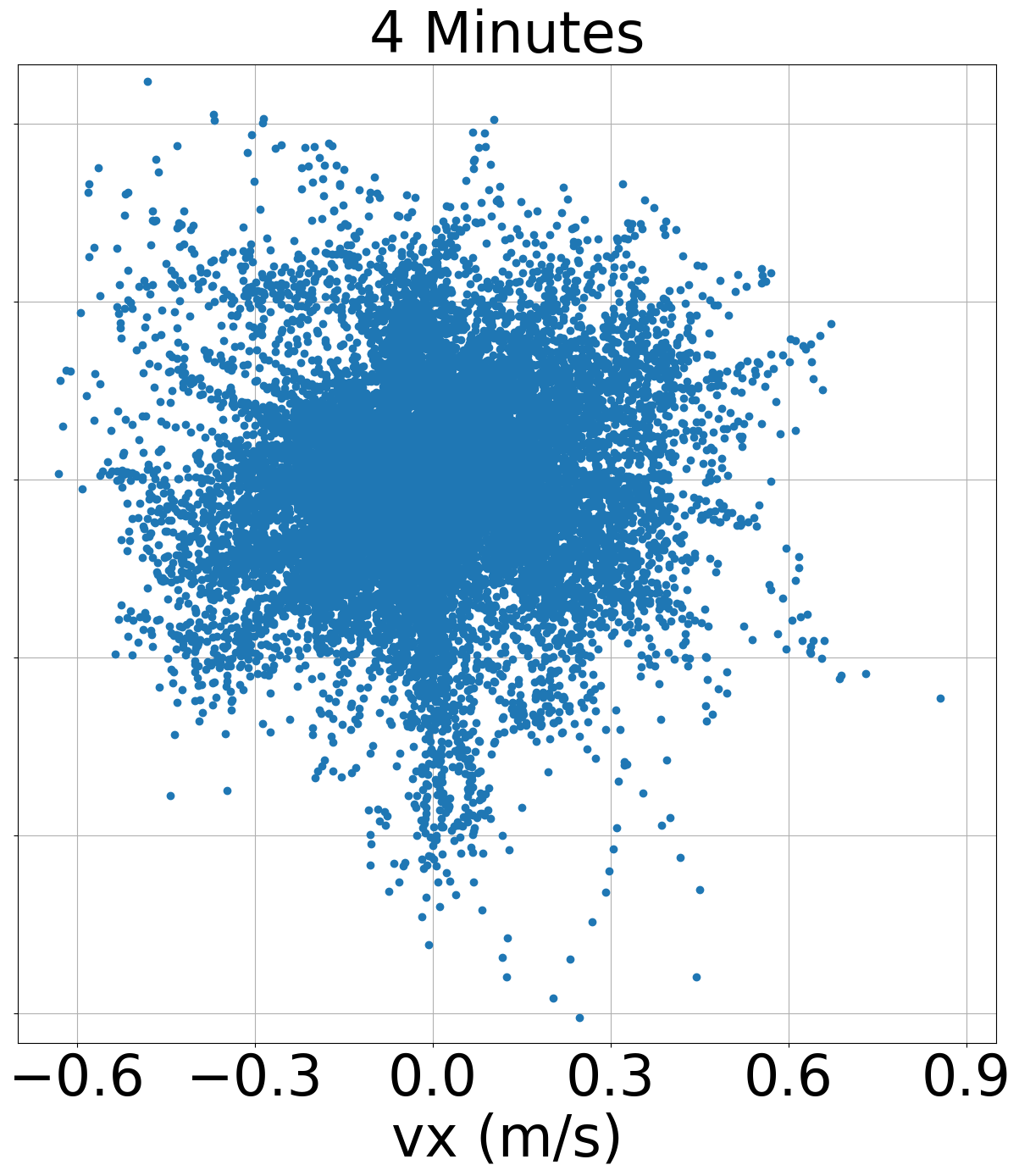}
    \end{subfigure}
    \begin{subfigure}[b]{0.18\textwidth}
        \centering
        \includegraphics[width=\textwidth]{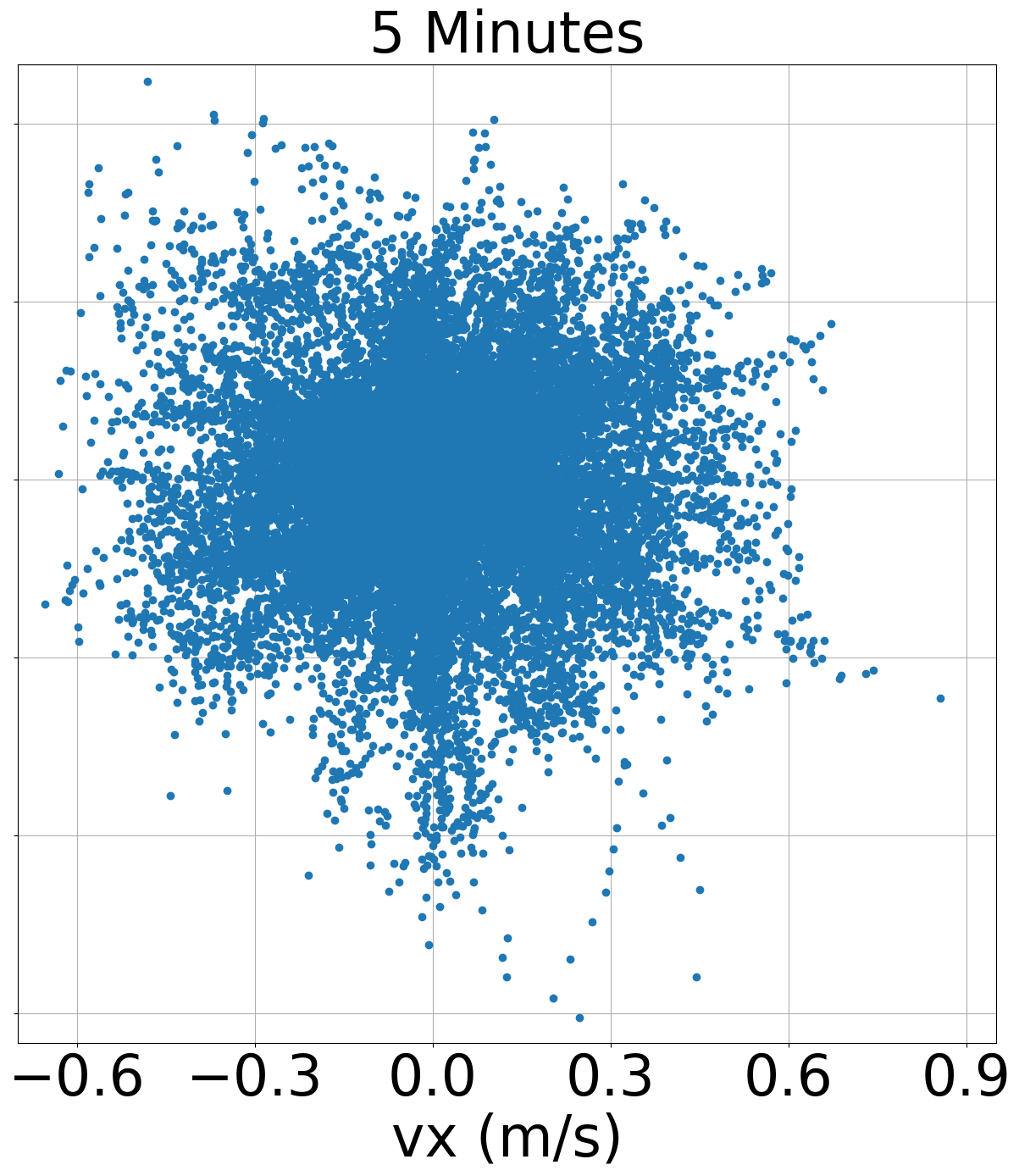}
    \end{subfigure}
    \caption{The distribution of the puck's velocities (m/s) in the initial dataset.}
    \label{fig:initial_dataset_velocity}
\end{figure}

We use the same initial dataset for all experiments on the Soft Stewart Platform. The distribution of samples are shown in Fig.~\ref{fig:initial_dataset_position} and~\ref{fig:initial_dataset_velocity}.

\end{document}